\documentclass[journal]{IEEEtran}

\usepackage[utf8]{inputenc} 
\usepackage[T1]{fontenc}    
\usepackage{hyperref}       
\usepackage{url}            
\usepackage{booktabs}       
\usepackage{amsfonts}       
\usepackage{amsmath}
\usepackage{microtype}      
\usepackage{graphicx}
\graphicspath{ {images/} }
\usepackage{multirow}
\usepackage{subcaption}
\usepackage{tabularx}

\usepackage{xcolor}

\newcommand{\bL}{\mathbf{L}}
\newcommand{\by}{\mathbf{y}}
\newcommand{\bx}{\mathbf{x}}

\newcommand{\bQ}{\mathbf{Q}}
\newcommand{\bX}{\mathbf{X}}
\newcommand{\bY}{\mathbf{Y}}
\newcommand{\bI}{\mathbf{I}}
\newcommand{\bT}{\mathbf{T}}
\newcommand{\bF}{\mathbf{F}}
\newcommand{\bB}{\mathbf{B}}

\newcommand{\bW}{\mathbf{W}}

\newcommand{\bJ}{\mathbf{J}}
\newcommand{\bA}{\mathbf{A}}
\newcommand{\bt}{\mathbf{t}}
\newcommand{\bd}{\mathbf{d}}
\newcommand{\balpha}{\boldsymbol{\alpha}}
\newcommand{\bSigma}{\boldsymbol{\Sigma}}
\newcommand{\btheta}{\boldsymbol{\theta}}
\newcommand{\bmu}{\boldsymbol{\mu}}
\newcommand{\smallpar}[1]{\textbf{#1}\hspace{0.2cm}}

\newcommand{\picm}[1]{\includegraphics[width=0.096\textwidth]{matting/#1}\hspace{0.1cm}}
\newcommand{\picsone}[1]{\includegraphics[width=0.145\textwidth]{segmentation/beta_1000/#1}\hspace{0.1cm}}
\newcommand{\picstwo}[1]{\includegraphics[width=0.14\textwidth]{segmentation/beta_100/#1}\hspace{0.1cm}}
\newcommand{\picd}[1]{\includegraphics[width=0.12\textwidth]{dehazing/#1}\hspace{0.1cm}}
\newcommand{\picdr}[1]{\includegraphics[width=0.12\textwidth]{dehazingreal/#1}\hspace{0.1cm}}

\begin{document}


\title{Deep-Energy: Unsupervised Training of \\Deep Neural Networks\\
}

\author{Alona~Golts,
        Daniel~Freedman,
        and~Michael~Elad,~\IEEEmembership{Fellow}
\thanks{A. Golts and M.Elad are from the Department
of Computer Science, Technion Institute of Technology, Technion City, Haifa 32000, Israel,
corresponding e-mails: salonaz@cs.technion.ac.il, elad@cs.technion.ac.il.
D. Freedman is from Google Research, Haifa, Israel, email: danielfreedman@google.com.}
}

\markboth{IEEE TRANSACTIONS ON IMAGE PROCESSING,~Vol.~X, No.~Y, October~2018}%
{Shell \MakeLowercase{\textit{et al.}}: Bare Demo of IEEEtran.cls for IEEE Journals}
%


\maketitle

\begin{abstract}
The success of deep learning has been due, in no small part, to the availability of large annotated datasets. Thus, a major bottleneck in current learning pipelines is the time-consuming human annotation of data. In scenarios where such input-output pairs cannot be collected, simulation is often used instead, leading to a domain-shift between synthesized and real-world data. This work offers an unsupervised alternative that relies on the availability of task-specific energy functions, replacing the generic supervised loss. Such energy functions are assumed to lead to the desired label as their minimizer given the input. The proposed approach, termed \emph{Deep-Energy}, trains a Deep Neural Network (DNN) to approximate this minimization for any chosen input. Once trained, a simple and fast feed-forward computation provides the inferred label. This approach allows us to perform unsupervised training of DNNs with real-world inputs only, and without the need for manually-annotated labels, nor synthetically created data. \emph{Deep-Energy} is demonstrated in this paper on three different tasks -- seeded segmentation, image matting and single image dehazing -- exposing its generality and wide applicability. Our experiments show that the solution provided by the network is often much better in quality than the one obtained by a direct minimization of the energy function, suggesting an added regularization property in our scheme.
\end{abstract}



\section{Introduction}

Deep learning \cite{deep_learning} has enjoyed a remarkable success in a wide range of fields, including speech recognition \cite{speech_recognition}, image processing \cite{fast_image_processing}, computer vision \cite{alexnet} and natural language processing \cite{NLP}. 
The predominant way of training deep neural networks (DNNs) is supervised, involving three major components: (1) Composition of the data into labeled input and output pairs; (2) Design of the architecture, and (3) Choosing the loss function to guide the training. While the first two require advanced domain knowledge specific to the task at hand, the choice of the loss function is mostly generic and application-independent. Common loss functions include $L_2$, $L_1$, cross-entropy and perceptual loss, all measuring the proximity between the network prediction and the ground truth output. For the learning process to succeed, many (thousands to millions) of input- and corresponding labeled output pairs are required.

While collection of large-scale annotated datasets is tolerable in classification tasks, the picture changes drastically in semantic segmentation, where pixel-wise labeling of each image\footnote{In Pascal VOC 2012 this task takes $\sim 4$ minutes per image.} is time consuming \cite{whats_the_point}, implying an impossible amount of work for gathering the training data. This bottleneck is even worse in medical imaging, where data is often of higher dimensions, and the annotators must be experienced radiologists \cite{medical_1,medical_2}. In other applications, e.g., single image dehazing, the collection of clear and hazy images of the exact same scene is generally impossible. The common practice in such cases is simulation of input-output pairs \cite{CAP,mscnn,dehazenet,aodnet,GFN}. Apart from requiring non-trivial domain knowledge, such a simulation can potentially inject errors into the learning process, creating a domain-shift when treating real-world data. 

All this brings us to the main question this paper addresses: can we train DNNs while avoiding the need for labels, bypassing the above-described bottleneck? In order to answer this question positively, we recall a classic and successful strategy in handling many computer vision and image processing tasks -- \emph{energy minimization}. In this pre-deep-learning era approach, a cost function over the unknown output is formulated, reflecting its desired relationship with the input, along with other penalties characterizing the desired result.  The ``inference'', i.e., getting the desired output for a given input, is obtained by minimizing this energy function. For example, in image denoising one such a cost function could force the output to resemble the input, while also being piece-wise smooth. Inference in this case will result in a smoothed, cartoon-like version of the input. The idea we promote in this work is to rely on such energy functions in order to avoid having explicit labels. However, where could we find appropriate energy functions? Fortunately, the computer vision and image processing literature is replete with examples of such functions for solving a wide variety of problems, such as depth from stereo \cite{energy_stereo}, super-resolution \cite{energy_sr}, single-image dehazing \cite{energy_dehazing}, optical flow estimation \cite{energy_optic}, and many more. 

Note that while in data-driven methods training is usually slow while inference is quite fast, in energy minimization the opposite is correct. There is no training involved, but inference requires a tedious (and possibly iterative) minimization of the energy function for each input signal, posing a difficulty in real-time applications. Additionally, any newly formulated energy function must be accompanied with a tailored optimization scheme. Indeed, in order to ease the optimization and use standard tools, often times the energy function is over-simplified while sacrificing output quality. 

Our proposed approach, termed \emph{Deep Energy}, offers an unsupervised training of DNNs by a direct minimization of a well-chosen energy function, suited for the task at hand. Specifically, we enforce the output of a DNN to minimize a task-specific energy function, when fed with the corresponding input. We do so by optimizing the parameters of the network using SGD (Stochastic Gradient Descent) and back-propagation, such that, averaged over all examples in the training set, the energy function is minimized. Since these energy functions are unsupervised (i.e. do not assume the knowledge of the output), this removes entirely the dependency on manually-annotated or synthetic input-output pairs.

As this work relies on the classic energy minimization approach and bridges it to the more recent data-driven methods, a natural question to ask is: Why bother learning at all? What is the benefit in the proposed scheme when compared to a direct optimization of the energy at inference time? The answer has several facets to it. The proposed approach replaces the minimization of the energy function by an approximation obtained using DNN, and as a consequence (i) we are free from devising a tailored optimization algorithm for the minimization task; (ii) we can handle more challenging energy functions, which are considered as hard to solve using existing tools; (iii) the inference itself, once the network has finished learning, is computed by an almost instant forward-pass operation, typically orders of magnitude faster than an explicit optimization; and most importantly, (iv) we show that the incorporated DNN induces an effective regularization over the analytic solver, improving its performance. Indeed, the last point is closely related to the main idea appearing in \cite{deep_image_prior}. To summarize, this work provides the following contributions:
\begin{enumerate}
	\item We suggest incorporating domain-knowledge into loss functions by a new methodology of unsupervised training of DNNs through task-specific energy functions.
	\item We demonstrate this concept by training a single network architecture to preform three different tasks -- seeded segmentation, image matting and single image dehazing -- each relying on its appropriate task-specific energy function.
	\item We provide a clear comparison between the optimization-based analytical solver and the network's approximated solution, showing that our method achieves an additional effective regularization, which improves the eventual solution in terms of both quality and speed.
\end{enumerate}

\noindent The remainder of this paper is structured as follows: Section \ref{s:related_work} reviews previous related work; Section \ref{s:method} introduces our \emph{Deep-Energy} approach and provides three different energy functions for treating seeded segmentation (\ref{ss:segmentation}), image matting (\ref{ss:matting}) and single image dehazing (\ref{ss:dehazing}). Section \ref{s:results} discusses our experimental environment and reports quantitative and qualitative results. Section \ref{s:conclusion} concludes this paper, and Appendix \ref{s:appendix} offers additional technical details.


\section{Related Work}\label{s:related_work}

\noindent \smallpar{Our General Approach:} 
Our approach is inspired by recent work on style transfer \cite{style_transfer}, which creates a stylized image by optimizing a two-term energy function. These two terms ensure perceptual loss proximity between the input image and its stylized output version, and a similar proximity of the second order statistics between the style and output images. Instead of repeated optimization of the energy function per each input image, the works in \cite{FF_style_transfer,FF_texture} train a feed-forward DNN to produce the stylized image by a near-instant forward-pass operation. They do so in an unsupervised manner by training the network to minimize the energy function directly. 

There is a growing body of work suggesting training DNNs using energy functions \cite{unsupervised_optical_flow,unsupervised_face_reconstruction,unsupervised_smoothing,normalized_cut_energy}.
In \cite{unsupervised_optical_flow}, a DNN is trained by minimizing the large-displacement photometric consistency for optical flow estimation. The work reported in \cite{unsupervised_face_reconstruction} refines coarse $3$D face models by training with an SfS (Shape-from-Shading) energy function. In \cite{unsupervised_smoothing}, a DNN is taught to perform image smoothing using an unsupervised energy-based loss function. Finally, the work most resembling ours, \cite{normalized_cut_energy}, performs semantic segmentation by combining a weakly-supervised cross-entropy term with an energy-based normalized-cuts regularization. While these works focus on achieving SOTA (State-Of-The-Art) in specific applications, we inspect energy-minimization training from a broader and holistic point of view. We formulate energy-based learning for three different applications and show the extra regularization obtained, along with the other benefits, arising from this approach.

Speeding-up image-processing operators, as reported in \cite{fast_image_processing}, bears some similarity to our approach as well. In \cite{fast_image_processing}, an analytical or numerical solver of an energy function is invoked thousands of times to create input and output pairs for supervised learning of a DNN. Once training of the DNN is over, evaluation can be made via an almost instant forward-pass operation, instead of an explicit optimization for each image. As opposed to \cite{fast_image_processing}, we skip the prolonged optimization for creating input-output pairs, and directly minimize the energy function during training. Doing so achieves effective regularization which actually improves the results of the analytical solver.

\noindent \smallpar{Past Work on Seeded Segmentation: } 
Seeded segmentation has enjoyed a fair amount of attention with the introduction of the popular Graph-Cuts \cite{classical_seg_1,classical_seg_2} and Random-Walker \cite{random_walker} algorithms. With the emergence of deep learning and fully annotated datasets, such as Pascal VOC 2012 \cite{pascal} and Imagenet \cite{imagenet}, the focus has shifted to automatic semantic segmentation \cite{supervised_segmentation}. Recent attempts to alleviate the burden of pixel-wise annotation have incorporated ``weak-supervision'' in the form of image-level labels \cite{weakly_darrell,seed_expand,stc}, bounding-boxes \cite{weaklysup,boxsup}, scribbles \cite{scribblesup} and points \cite{whats_the_point}. 

These approaches differ from ours in two important aspects: (1) They perform weakly-supervised semantic segmentation, i.e., the additional supervision is given only at training time, whereas our energy-based seeded segmentation receives seeds both at training and at test time; and (2) Most of these works use an external analytical solver for either pre-processing of input-output pairs, or post-processing to refine the result of the network. We circumvent the use of an external analytical solver and train end-to-end on the energy function directly, reducing time and effort in both training and evaluation.

\noindent \smallpar{Past Work on Image Matting: }  
The ill-posed problem of image matting commonly requires additional user assistance in the form of seeds or trimaps. A trimap is a rough segmentation, dividing the input image to three sections: constant foreground and background pixels, and unknown pixels whose opacity is to be determined. The two leading approaches of tackling image matting are prior-based and learning-based. Prior-based methods \cite{closed_form_matting,classic_matting_1,classic_matting_2,classic_matting_3,classic_matting_4,classic_matting_5} mitigate the ambiguity in image matting by formulating energy functions whose solution entails a non-trivial optimization. Conversely, learning-based methods \cite{deep_matting,deep_matting_1,deep_matting_2} compose input-output pairs for supervised training of DNNs. However, obtaining the ground-truth pixel-wise opacity of a furry/semi-opaque object in the wild is generally infeasible. A common alternative is training on simulated images, either created with over-simplified backgrounds \cite{alpha_matting_com}, or as compositions of non-related foregrounds and backgrounds \cite{deep_matting}. This approach may create a domain-shift in treating challenging real-world images. Our method avoids utilizing the ground truth during training altogether, circumventing this issue entirely.

\noindent \smallpar{Past Work on Single Image Dehazing: }
Prior to the emergence of deep learning solutions, single image dehazing has been handled as an energy minimization task \cite{contrast_tan,energy_dehazing,BCCR,color_lines,NLD}. This implied choosing a clever prior and incorporating it into an energy function, followed by an execution of a carefully selected optimization scheme. However, in addition to performing repeated optimization for each image, the returned results were often characterized as having non-physical colors and increased saturation and contrast. In recent years, there has been a general shift towards supervised data-driven and deep-learning-based solutions \cite{CAP,mscnn,dehazenet,aodnet,GFN}. Generally, it is impossible to capture clear and hazy images sharing the exact same time, scene and lighting conditions. Thus, supervised methods resort to simulating input and output pairs, while relying on the haze formation model \cite{haze_model}. Hazy images can be reliably synthesized using clear-day images and their corresponding depth maps. However, outdoor depth maps are costly and highly inaccurate \cite{reside}; hence, indoor depth images are frequently used instead. This leads to an inherent domain shift when treating outdoor hazy images which are the final target of dehazing algorithms. Our proposed scheme evades this issue by training on real-world outdoor hazy images only. 


\section{Proposed Approach}\label{s:method}

In this section we present the \emph{Deep Energy} scheme, first describing it in general terms, and then discussing three different image processing tasks that harness it.

\subsection{\emph{Deep Energy}: General Scheme} \label{ss:general}

Our interpretation of an ``Energy Function'' is simply a cost function $E(\bx,\by)$ for the given input $\bx$ and the unknown output $\by$, describing a desired behaviour of the output. For example, the output should be required to be close to the input, while being piece-wise-constant. These two conditions can be explicitly enforced by formulating a specific energy function of the form
\begin{equation}
E(\bx,\by) = \by^T \bL \by + (\by-\bx)^T(\by-\bx),
\end{equation} 
where $\bL$ is a smoothness-enforcing Laplacian matrix. Considering a general energy function $E(\bx,\by)$, the optimal output given an input $\bx$ can be derived by
\begin{equation}
\label{eq:inverse}
\hat{\by} = \arg \min_{\by} E(\bx,\by).
\end{equation}
The process of minimizing the energy function, called ``inference'', might be computationally exhaustive, depending on the specific energy function, the dimensions involved, and the chosen optimization method. In its core, the reliance on an energy function poses an \textit{unsupervised} approach, as it does not require any learning, and the optimization is conducted for each signal $\bx$ individually. The above line of reasoning has been quite popular over the past three decades, and numerous energy functions have been proposed and leveraged for tackling different tasks in the fields of computer vision and signal and image processing.  

Instead of repeatedly solving the energy function for each input instance $\bx^m$, we suggest training a DNN to solve the problem posed in Equation (\ref{eq:inverse}). We do so by directly minimizing an unsupervised energy loss during training. Given an unlabelled dataset $\mathcal{X}$, $\{\bx^n\}_{m=1}^M$, we propose representing the output of the energy function as the prediction of a DNN $\by_p(\bx,\theta)$ where `p' stands for ``prediction'', $\theta$ are the network parameters, and the network's input is $\bx$. To tune the network parameters, we minimize the following energy-based equivalent of the empirical loss over the training set:
\begin{equation}
\min_{\btheta} \frac{1}{M} \sum_{m=1}^M E(\bx^m, \by_p(\bx^m; \btheta)).
\label{eq:deep_energy}
\end{equation}
Thus we learn the network such that, averaged over all examples, the energy function is minimized. This promotes a desired behavior of the output without enforcing closeness to ground-truth labels. We call this strategy \emph{Deep Energy} -- unsupervised learning by minimization of a task-specific energy function. 
The proposed scheme can serve any application, as long as it can be posed in terms of an informative energy function, and for which there is difficulty in collecting large-scale data for supervised learning. Examples for such tasks are optical flow estimation \cite{optical_flow}, single image depth estimation \cite{depth_estimate}, image retargetting \cite{retargetting}, Retinex \cite{retinex}, and more. In the remaining part of this section we describe three applications in which we have incorporated \emph{Deep Energy}: seeded segmentation, image matting and single image dehazing.


\subsection{Application 1: Seeded Segmentation} \label{ss:segmentation}

In seeded segmentation, the input consists of an image to be segmented, and a sparse set of user-provided labelled seeds, indicating the initial location of each object in the image; the output is a pixelwise segmentation map. Seeded segmentation has enjoyed popularity within the medical imaging community \cite{medical_seeded_seg}, allowing the user a degree of control over the final segmentation result. In recent years, with the emergence of deep learning, seeded segmentation has been treated from a ``weakly-supervised'' perspective. In this regime, the user-provided seeds are only given at training, whereas the segmentation is performed automatically at test time, without user assistance. 

The majority of ``weakly-supervised'' methods \cite{weakly_darrell,weaklysup,scribblesup,whats_the_point} use external analytic solvers to extract input-output pairs for supervised training. We, on the other hand, propose to use the \emph{Deep Energy} scheme, thereby avoiding the need for labeling altogether. For this purpose, we use the energy function of the well-known Random Walker algorithm by Grady \textit{et al.} \cite{random_walker}, in which the image is represented as a weighted undirected graph $G = (V,\mathcal{E},W)$. The vertices $V$ are the pixels, and the edges $\mathcal{E}$ correspond to their 4-neighborhoods. The weight of an edge $e_{ij}$ is given by $w_{ij} = \exp\{-\beta \|\bI_i - \bI_j\|^2\}$, where $\bI_i,\bI_j$ are the RGB values of the pixels $i$ and $j$, and $\beta$ is a global scaling parameter.


Random walker models each pixel as conducting a random walk over the above described graph. The probability of a given pixel reaching each and every seed/class in the image is also the final score of that class. Formally, let $\by^l \in \mathbb{R}^N$ be the pixelwise probability of belonging to class $l$, where $N$ denotes the number of pixels in the image. 
Denote $\bx^l \in \mathbb{R}^N$ as the ``seed image'' of class $l$ whose elements are $0$ or $1$, indicating absence or presence of a seed respectively. The vectorized versions of the initial seed and the final probabilities over $L$ classes are denoted by $\bY,\bX \in \mathbb{R}^{N\times L}$, where each column corresponds to a separate class. The random walker probabilities, $\by^l,~ l \in [1,L]$, can be computed by minimizing the following energy function\footnote{We use a soft version of random walker \cite{random_walker}, instead of the original formulation, which uses a hard constraint on the seed locations.} \cite{random_walker}:
\begin{eqnarray}\label{eq:Energy}
E(\bX,\bY) & = & \sum_l (\by^l)^T \bL \by^l + \lambda \sum_l (\by^l-\bx^l)^T \bQ (\by^l-\bx^l)  \nonumber \\
		   & = & tr \left\{ \bY^T \bL \bY + \lambda(\bY-\bX)^T \bQ (\bY-\bX)\right\},
\end{eqnarray}
where $\bQ = \text{diag}(\sum_l \bx^l) \in \mathbb{R}^{N \times N}$ is a matrix whose diagonal elements indicate the presence of a seed of any class at a given pixel, and $\bL$ is the Laplacian of the graph given as
\begin{equation}\label{eq:Laplacian}
\bL_{ij} =
\begin{cases}
d_i = \sum_j w_{ij}         & \text{if } i=j, \\
-w_{ij}         & \text{if } v_i \text{ and } v_j \text{ are adjacent nodes,} \\
0               & \text{otherwise}.
\end{cases}
\end{equation} 
The first term in Equation (\ref{eq:Energy}) is a ``smoothness term'', ensuring that adjacent pixels with similar colors have similar output probabilities. Conversely, the weights of adjacent pixels in discontinuous regions should be close to zero, allowing for different probabilities in the output. The second is a ``data term'', encouraging fidelity to the input seeds. The final loss function for training our DNN is a tensor-friendly version of Equation (\ref{eq:Energy}), explained in details in Appendix \ref{s:appendix}. 

Returning to our  \emph{Deep Energy} paradigm, our aim is to use it in order to put forward an alternative and faster way for handling this segmentation task. Given a set of input images and their seeds, $\{\bI^m,\bX^m\}_{m=1}^M$, our loss module computes the corresponding matrices $\{\bL^m, \bQ^m\}_{m=1}^M$, and optimizes for the parameters $\theta$ of the network $\bY_p(\bI^m,\bX^m;\theta)$, such that its prediction over all training examples $\{\bY^m\}_{m=1}^M$ minimizes the expected energy function described above. After training has finalized, the network has learned to approximate the output probability $\bY^{test}$, corresponding to a new set of image and seeds $\{\bI^{test},\bX^{test}\}$, done via a simple forward-pass operation.


\subsection{Application 2: Image Matting} \label{ss:matting}

In image matting, an object is extracted from its background by determining the opacity and color of each pixel in the foreground. The input is an image, assumed to be a composite of foreground and background images, and the output is an alpha matte, indicating the opacity of the foreground versus the background in each pixel. The following is known as the ``matting equation'',
\begin{equation}\label{eq:compositing}
\bI_i = \alpha_i \bF_i + (1-\alpha_i) \bB_i, \quad \forall i \in \bI,
\end{equation}
where $\bI \in \mathbb{R}^{N \times 3}$ is the input RGB image, $\bF,\bB \in \mathbb{R}^{N \times 3}$ are the unknown foreground and background images, and $\balpha \in \mathbb{R}^{N} \in \left[0...1\right]$ is the alpha matte. Note that recovering the alpha matte from a single input image is extremely ill-posed, since seven quantities per each pixel must be deduced (the RGB values of $\bF_i,\bB_i$ and the alpha matte $\alpha_i$).

The energy function we use for this task is the closed-form matting by Levin \textit{et al.} \cite{closed_form_matting}. By incorporating minimal user interaction in the form of seeds and assuming that the background and foreground are locally smooth, \cite{closed_form_matting} manages to eliminate the dependency on $\bB,\bF$ and obtain a closed-form solution for $\balpha$. Computing the solution of $\balpha$ is obtained by minimizing 
\begin{equation}\label{eq:matting_energy}
  E(\bx, \balpha) = \balpha ^T \bL \balpha + \lambda (\balpha - \bx)^T \bQ (\balpha - \bx),
\end{equation}
where $\bQ \in \mathbb{R}^{N \times N}$ is a diagonal matrix whose nonzero elements indicate the presence of a foreground or background seed. The vector $\bx \in \mathbb{R}^{N}$ is the seed image with $'1'$s in respective foreground seed locations and $'0'$s elsewhere. The matrix $\bL$ is a special Laplacian-like matrix \cite{closed_form_matting} given by
\begin{equation}\label{eq:matting_laplacian_color}
\bL_{ij} = \sum_{n | (i,j) \in p_n} (\delta_{ij} - w_{ij}^n),
\end{equation}
where
\begin{equation}\nonumber
w_{ij}^n = \frac{1}{|p_n|} (1 + \left(\bI_i - \bmu_n\right)^T(\bSigma_n + \frac{\varepsilon}{|p_n|}\textbf{I})^{-1}\left(\bI_j - \bmu_n\right)).
\end{equation}
In the above, $i,j$ are pixels within the patch $p_n$, centered at the pixel $n$; $\bmu_n \in \mathbb{R}^{3 \times 1},\bSigma_n \in \mathbb{R}^{3 \times 3}$ are the mean and covariance of the patch of size $|p_n|$; $\textbf{I} \in \mathbb{R}^{3 \times 3}$ is the identity matrix, and $\varepsilon$ provides an additional control over the smoothness. The reformulation of Equation (\ref{eq:matting_energy}) as a loss function for training is detailed in Appendix \ref{s:appendix}.

We should note the clear similarity between this formulation and the random walker used in the seeded segmentation. Still, there are two main differences between the two: (i) The matting problem recovers a single image layer, whereas the segmentation produces $L$ layers; and (ii) The Laplacian $\bL$ matrices are formed very differently. 

As for the deployment of \emph{Deep Energy}, during training, pairs of input images and their seeds $\{\bI^m,\bx^m\}_{m=1}^M$ are fed to the network, and the \emph{Deep Energy} module computes the intermediate $\{\bL^m,\bQ^m\}_{m=1}^M$ matrices for the corresponding matting and data terms. The SGD optimization tunes the network parameters such that the predictions $\{\balpha^m\}_{m=1}^M$ minimize the average loss function in Equation (\ref{eq:matting_energy}). At inference, a new $\balpha^{test}$ is computed for a fresh set  $\{\bI^{test},\bx^{test}\}$ by a simple forward pass computation.


\subsection{Application 3: Single Image Dehazing} \label{ss:dehazing}

While seeded segmentation and image matting contain some user intervention, the third application we consider, single image dehazing, is completely unsupervised. Given a hazy input image, our goal is to lift the haze and reveal the hidden details in the image. This is a vital preliminary stage in many computer vision pipelines, including autonomous car navigation and object detection.

The common haze formation model \cite{haze_model} postulates a hazy image $\bI \in \mathbb{R}^{N \times 3}$ as a linear combination of a clear-day scene radiance image $\bJ \in \mathbb{R}^{N \times 3}$ and a constant atmospheric component $\bA \in \mathbb{R}^{3 \times 1}$, called the ``airlight''. The relation between these two is governed by a transmission map $\bt \in \mathbb{R}^{N}$, dependent on the pixelwise depth $\bd \in \mathbb{R}^{N}$ from the camera:
\begin{equation}\label{eq:haze_equation}
\begin{split}
\bI_i &= t_i \bJ_i + (1-t_i)\bA,   \quad \forall i \in \bI\\    
t_i &= \exp^{-\beta d_i}, \\
\end{split}
\end{equation}
where $\beta$ is a  scattering coefficient, controlling haze thickness. This model induces an under-determined set of $3N$ equations for the known pixels of $\bI$ and $4N+3$ unknowns for $\bJ,\bt,\bA$.
 
We adopt the energy function of the well-known ``Dark Channel Prior'' (DCP) \cite{energy_dehazing} by He \textit{et al.}, who resolved the ambiguity of the haze formation model by assuming a specific prior. DCP is based on a statistical property of natural outdoor images (excluding sky regions), stating that in small patches, the darkest pixel of the patch across all RGB channels tends to zero. This is due to shades and naturally dark and monochromatic objects. The ``Dark Image'' is defined as a minimum filter over small patches across RGB channels. In case of natural outdoor images, this dark image is mostly zero, 
\begin{equation}
J_i^\text{dark} = \min_{c \in \{r,g,b\}}({\min_{k \in \Omega(i)}}(\bJ^c_k)) \rightarrow 0,  \quad \forall i \in \bI
\end{equation}
where $\Omega(i)$ is a patch around pixel i, and $c$ are the RGB channels. Envoking the DCP on both $\bI$ and $\bJ$ in the haze formation model and assuming a constant transmission within a small patch, results in the following coarse transmission map: 
\begin{equation}\label{eq:dark_channel_prior}
\tilde{t_i} = 1 -  \omega \cdot \min_{c \in \{r,g,b\}} \left(\min_{k \in \Omega(i)} \frac{\bI^c_k}{A^c}\right), \quad \forall i \in \bI
\end{equation}
where the constant $\omega$ injects a small amount of haze for better visual perception. In \cite{energy_dehazing}, the resulting block-artifact-filled transmission map $\tilde{\bt}$ is smoothed with the same image matting technique in Section \ref{ss:matting} by Levin \textit{et al.} \cite{closed_form_matting}. The energy function that provides the refined version of the transmission map $\bt$ is given by
\begin{equation}\label{eq:dehazing_energy}
E(\bt,\tilde{\bt}) = \bt^T \bL \bt + \lambda(\bt-\tilde{\bt})^T (\bt-\tilde{\bt}),
\end{equation}
where $\bL$ is the matting Laplacian from Equation. (\ref{eq:matting_laplacian_color}), encapsulating the inter-pixel relations in the input hazy image $\bI$. 
As opposed to image matting, there are no user-provided seeds and the data term enforces closeness to all pixels in the coarse transmittance map $\tilde{\bt}$. The detailed implementation of Equation (\ref{eq:dehazing_energy}) as a loss function for training is given in Appendix \ref{s:appendix}. 

During training, the network is given the hazy images $\{\bI^m\}_{m=1}^M$ and \emph{Deep Energy} computes the intermediate $\{\tilde{\bt}^m,\bA^m\}_{m=1}^M$ and the final smoothing and data terms. At test time, a new hazy image enters the learned network that computes the predicted transmission map $\bt_{\theta}$. Given this predicted transmission, the scene radiance $\bJ$ can be computed using the haze formation model
\begin{equation}
\bJ = \frac{\bI-\bA}{\max(\bt_{\theta},t_0)} + \bA,
\end{equation}
where $t_0$ is a small constant threshold, used to avoid division by zero. The only missing quantity is the airlight $\bA$. We follow the heuristic in \cite{energy_dehazing} of inspecting the $0.1\%$ brightest pixels in the dark channel image of $\bI$. Out of these locations, the RGB value of the brightest pixel in $\bI$ is chosen as the airlight $\bA$.


\section{Results} \label{s:results}

We now present the results of our \emph{Deep Energy} approach on the three applications described above: the two weakly-supervised tasks of seeded segmentation and image matting and the fully unsupervised implementation of single image dehazing. Our results clearly show that through energy-based training, the network is able to mimic the results of the original classical solver, and even outperform it in terms of application-specific criteria. 

\begin{figure*}[t!]
\centering
\includegraphics[width=0.87\textwidth]{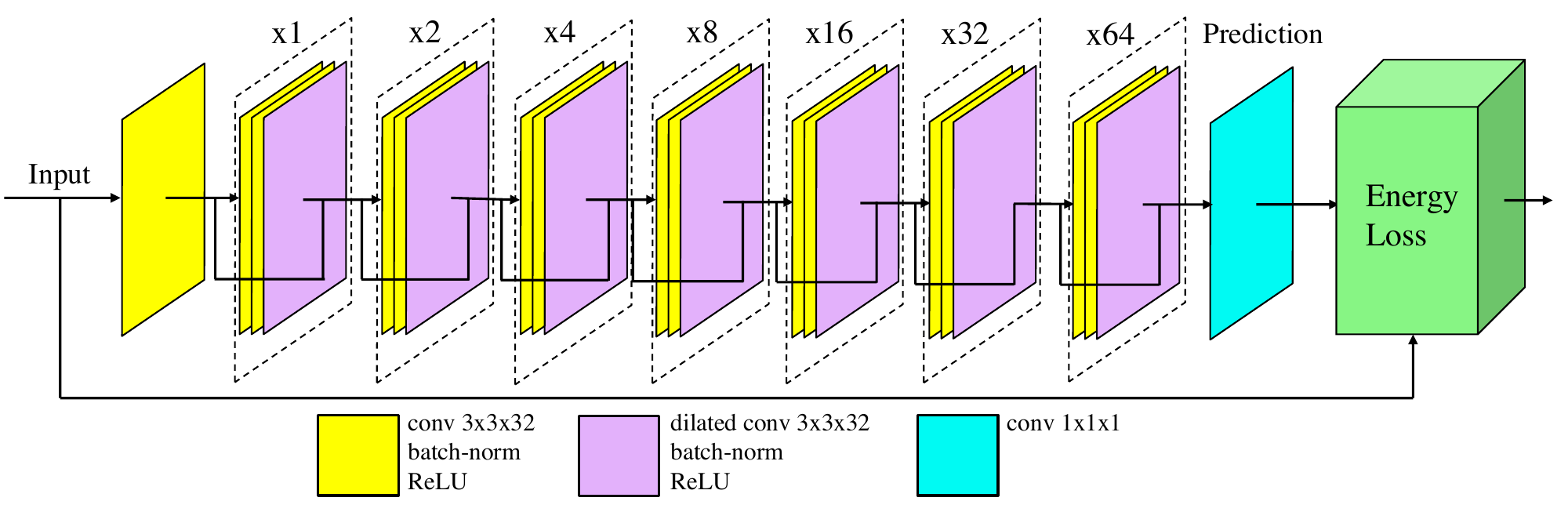}
\caption{System architecture. At input, the network receives either regular images or images concatenated with seeds, and outputs a prediction. Our network is a series of ``Dilated Residual Blocks'' in which the spatial dimensions remain intact, but the receptive field increases due to dilated convolutions with increasing dilation factor. Additional Resnet-style skip-connections between the input and output of each block are added for improved gradient flow.
The \emph{Deep Energy} loss receives only the input and prediction of the network, without relying on fully-annotated ground-truth labels.}
\label{fig:architecture}
\end{figure*}


\subsection{Architecture} \label{ss:architecture}

We use the same network structure for all three applications, as it was found to provide favorable results for various image-to-image applications. Our network architecture, shown in Figure \ref{fig:architecture}, is fully convolutional and based on the Context Aggregation Network (CAN), introduced in \cite{CAN}. The spatial dimension of the input, output and all intermediate layers remains the same, without using pooling or strided convolutions, preserving fine details in the image. To capture larger-scale semantic structures, we use dilated convolutions that also aggregate the sparse seed information in both seeded segmentation and image matting. These dilated convolutions have exponentially increasing dilation rates, this way enlarging the network receptive field.
We incorporate Resnet-style \cite{Resnet} skip connections to allow for additional gradient flow through the network layers and to facilitate the direct propagation of finer details in the image.

Specifically, our network is a cascade of ``Dilated Residual Blocks''. Each block contains two regular convolution layers, followed by a single dilated-convolution layer. The dilation factor is $2^d$, where $d$ is the block number, starting from dilation of $1$, then $2,4$, and so on, until $2^d$. All convolution layers, apart from the final output, are followed by batch normalization and ReLU nonlinearity, and have $3 \times 3$ filters with an output width of $32$. The final output layer is a linear $1 \times 1$ convolution of width equal to the size of the desired output. In seeded segmentation, there is an additional softmax layer to output the probabilities of Random Walker.
Finally, we add Resnet-style addition skip connections between the input and output of each block, as shown in Figure \ref{fig:architecture}. 


\subsection{Datasets} \label{ss:dataset}

\noindent {\bf Seeded Segmentation and Image Matting}:
We utilize the Pascal VOC 2012 dataset \cite{pascal}, augmented with extra annotations from \cite{pascal_aug}. We randomly split the original $10,582$ training images to $500$ for validation and parameter tuning, and the rest for training. We use the original $1,449$ images in Pascal `val' as test. We adopt the seed annotations supplied in \cite{scribblesup}. Since seeded segmentation is originally intended as user-assisted segmentation, we use the seeds during training and test time. To determine the quality of our solution during validation and test, we evaluate the mean-Intersection-over-Union (mIoU), averaged over all $21$ classes.

Image matting is originally a two-class task of separating a foreground object from its background. Thus, to construct two-class images out of the multi-class Pascal VOC, we create $L$ copies of the same $L$-class  natural image, with different seed locations from \cite{scribblesup} for each individual class. We repeat the same train-validation split and obtain two-class images, resulting in $14,772$ training and $735$ validation images.
Note that Pascal VOC was not originally intended for image matting. Two possible  alternatives are (i) Smaller datasets \cite{alpha_matting_com} that are less suitable for deep learning applications, and (ii) Carefully controlled simulated datasets \cite{deep_matting} that may not fully capture the behavior in the wild. Instead, we train on easy-to-obtain natural images and use the minimal seeds described above, rather than complex trimaps commonly used in image matting \cite{alpha_matting_com,deep_matting}. We evaluate the performance of the trained network on the $27$ test images in \textit{alphamatting.com}, accompanied with corresponding trimaps. We treat the constant background and foreground pixels as seeds; the missing gray pixels are completed by the algorithm. 

\noindent  {\bf Single Image Dehazing}: We use the RESIDE (REalistic Single Image DEhazing) dataset \cite{reside}, containing both real-world and simulated hazy images, accompanied with corresponding clear-day ground-truths. In general, collecting real-world pairs of clear and hazy images is impossible. Thus, learning-based methods resort to simulating hazy images with the haze formation model in Equation (\ref{eq:haze_equation}). Given a clear-day image and its corresponding depth map, one can generate multiple hazy images with varying amounts of haze thickness $\beta$ and airlight components $\bA$. Since outdoor depth maps are less accurate and much harder to acquire, the common practice is using indoor depth data and creating an indoor dehazing dataset. We, on the other hand, do not need the ground-truth outputs and can directly train on the RTTS -- RESIDE's collection of $4,322$ real-world outdoor hazy images. To evaluate the PSNR (Peak Signal-to-Noise Ratio) and SSIM (Structured-Similarity) values of our trained model during validation, we use a random subset of $500$ images from RESIDE's OTS dataset. Our test set is RESIDE's collection of $500$ outdoor images, called ``SOTS-outdoor'', and the smaller HSTS, consisting of $10$ outdoor images.


\subsection{Implementation Details} \label{ss:implementation}

\noindent {\bf Data Augmentation}: In image matting no data augmentation is performed and the single-class images of Pascal VOC are resized using bilinear interpolation to $128 \times 128$. In seeded segmentation we enlarge the train set by a factor of four to $40,328$ images. The first augmentation is a simple resize of the Pascal VOC images to $128 \times 128$. The second, third and fourth augmentations are obtained by a horizontal flipping, and random crop to a random-sized square (between $200$ pixels and the minimum dimension of the image), and random rotation by $0,~90,~180,~270$ degrees. All resulting augmentations are resized using bilinear interpolation to $128 \times 128$.
In Single image dehazing, we again enlarge the training set by a factor of four. The first set are simply the train images resized to $128 \times 128$. The other augmentations are horizontal flips, random crops to $256 \times 256$ or $512 \times 512$, and random rotations by $0,45,90,135$ degrees. Finally, the images are resized to $128 \times 128$. This creates a total of $17,288$ training images.

\noindent  {\bf Experimental Setup}: We implement our scheme in TensorFlow on a GTX Titan-X Nvidia GPU. In all applications we use the Adam \cite{Adam} optimizer for training, and draw the initial network weights from a Gaussian distribution $\mathcal{N} \sim (0,0.1)$. 
In seeded segmentation, the network consists of $7$ dilated residual blocks with zero padding, and a maximum dilation factor of $\times 64$. Although our network is fully-convolutional, na\"ive forward-pass of a large image results in an insufficient spread of the seeds. Instead, to evaluate a new larger-sized image, we follow \cite{normalized_cut_energy} and resize it (and its seed) to $128 \times 128$, pass it through the network, and resize the resulting segmentation or probability maps back to original size. We found that simple nearest neighbor interpolation of the segmentation result is accurate while being much faster than interpolating $21$ probability maps. 

Our training schedule operates as follows: we first initialize the learning rate to $0.01$ and decrease it by a factor of $\sqrt{2}$ every two sign changes of the derivative of the validation loss. We allow a cool-off period of two epochs, during which sign changes are ignored. Finally, our batch size is $13$. As to the parameters of the energy function, we found the effect of the parameter $\lambda$ to be negligible in comparison to $\beta$. Thus, we set $\lambda=1$ and tune only the $\beta$ parameter using the validation set. The best performance of the analytic solution is obtained for $\beta=100$ and the network's best solution corresponds to $\beta=1000$. For our best configuration of $\beta=1000$, we stop training after $25$ epochs, corresponding to $15$ hours.

In image matting, the network consists of $6$ dilated residual blocks with reflective padding. The learning rate schedule is similar to seeded segmentation and starts from an initial rate of $0.01$, decreases by $\sqrt{2}$ every $4$ sign changes of the derivative of the validation loss, and features a $2$ epoch cool-off period. The batch size is $10$ and the hyper-parameters of the energy function are as recommended in \cite{closed_form_matting}: the patch size is set to $3 \times 3$, and $\lambda=1,\varepsilon=10^{-5}$. Training lasts for $59$ epochs, corresponding to roughly $35$ hours. During validation, we perform a fully-convolutional forward-pass through the network, with the original input dimensions.

Finally, in single image dehazing the network consists of $6$ dilated residual blocks with zero padding. The learning rate starts from $3 \times 10^{-4}$, exponentially decreased by a factor of $0.96$ every $3$ epochs. The batch size is $24$, and the energy function hyper-parameters are adopted exactly from \cite{energy_dehazing}: $\lambda = 10^{-4}, \omega =0.95, t_0 =0.1, \epsilon = 10^{-6}$, the DCP patch size is $15 \times 15$, and the soft matting patch size is $3 \times 3$. Training is stopped after $30$ epochs, which are roughly $8$ hours.


\subsection{Quantitative Results} \label{ss:quantitative}

First, we wish to quantify the proximity of the trained network solution to the ``analytic solution'', obtained by directly minimizing the energy functions in Equations (\ref{eq:Energy}), (\ref{eq:matting_energy}), and (\ref{eq:dehazing_energy}). By plugging-in the solution of the network or the analytic solver back to the loss function, one can obtain a loss value. We measure this average loss value over the test set for both the analytic and the network solutions and report the results in Table \ref{tbl:results}. Additionally, we compare the overall quality of the solutions by task-specific criteria. In seeded segmentation, we provide the mIOU score over the Pascal VOC `val' dataset\footnote{We report the loss of the $128 \times 128$ images and the mIOU score of the fully-resized segmentation results of both the analytic and network solutions.}. In image matting, we report the MSE (Mean Square Error) and SAD (Sum Absolute Differences) scores of both solutions on the training set of \textit{alphamatting.com}. Finally, in single image dehazing we measure the PSNR and SSIM metrics on SOTS-outdoor and HSTS. These results are given in Table \ref{tbl:results} as well.

In seeded segmentation we show the performance for $\beta=100$ and $\beta=1000$, the best hyper-parameter values for the analytic and network solutions respectively. In both cases, the average loss value of the analytic solution is lower than that of the network. Clearly, \emph{Deep Energy} cannot reach the absolute minimum of the loss function with the chosen architecture\footnote{In fact, this is true for all three applications.}. Nonetheless, in terms of the actual quality of the obtained segmentation, represented by the mIOU metric, the network's solution is better. For $\beta=100$ it slightly outperforms the analytic solution, and for $\beta=1000$ it improves it considerably. In addition, while the analytic solution is sensitive to the choice of $\beta$ ($5\%$ difference in mIOU between the two configurations), our network is more robust. 

In image matting, the analytic solver outperforms the network solution. This result is expected since we made a compromise of training on Pascal VOC images, originally intended for the coarser task of semantic segmentation. The test set, however, consists of challenging furry and hairy objects, relevant to image matting. The ideal solution would be constructing a large-scale weakly-annotated image matting dataset, but it is out of the scope of this paper. That said, the network's solution is competitive due to its much improved run-time. 

Finally, in single image dehazing, we encounter an even stronger regularization ability as compared to the analytic solution, reflected in a $7$ dB increase in PSNR and a substantial increase in SSIM. We believe that this large gap is attributed to the relative weakness of the DCP energy in dealing with the sky regions in outdoor images. Our network is able to mitigate this difficulty by facing thousands of real-world hazy images and characterizing the general appearance of the sky.

Recall that our initial goal was teaching the network to approximate the minimizer of the energy function. However, by early-stopping the learning process, before reaching the absolute minimum, we often obtain effective regularization. This regularization may stem either from the network architecture, as shown in \cite{deep_image_prior}, or from the learning process, as shown in \cite{implicit_bias}. Further extensive exploration of this regularization is left for future work. 

\begin{table*}[t]
  \caption{Quantitative results of the \emph{Deep Energy} approach. Each cell describes (analytic/network) performance values, and the better is shown in Bold. In mIOU, SAD, PSNR and SSIM, higher is better, and in MSE lower is better.}
  \label{tbl:results}
  \centering
  \def\arraystretch{1.3}
  \resizebox{0.8\textwidth}{!}{
  \begin{tabular}{l|cc|cc|cc}
    \hline \hline
    & \multicolumn{2}{c|}{Seeded Segmentation} & \multicolumn{2}{c|}{Image Matting} & \multicolumn{2}{c}{Single Image Dehazing} \\
    \hline
    & $\beta=100$ & $\beta=1000$ & fine trimap & coarse trimap & HSTS & SOTS-outdoor\\
    \hline
    Loss & $\textbf{5.153}/8.918$ & $\textbf{0.669}/3.425$ & $\textbf{1.732}/4.141$ & -- & $\textbf{0.138}/0.910$ & $\textbf{0.158}/1.059$  \\
    mIOU [$\%$] & $71.99/\textbf{72.12}$ & $66.97/\textbf{73.76}$ & -- & -- & -- & --  \\
    MSE & -- & -- & $\textbf{0.029}/0.038$ & $\textbf{0.033}/0.047$ & -- & --  \\
    SAD & -- & -- & $\textbf{6,810}/9,990$ & $\textbf{10,512}/17,709$ & -- & --  \\
    PSNR [dB] & -- & -- & -- & -- & $15.96/\textbf{24.41}$ & $16.96/\textbf{24.07}$  \\
    SSIM & -- & -- & -- & -- & $0.877/\textbf{0.934}$ & $0.886/\textbf{0.933}$  \\
    Time [sec] & $0.657/\textbf{0.029}(\times 22)$ & $1.599/\textbf{0.029}(\times 56)$ & $21.635/\textbf{0.532} (\times 41)$ & $24.789/\textbf{0.524} (\times 47)$ & $28.018/\textbf{0.506}(\times 55)$ & $28.887/\textbf{0.582}(\times 50)$ \\
    \hline \hline
  \end{tabular}
  }
\end{table*}


\subsection{Runtime Comparison} \label{ss:runtime}

The analytic solver, as efficient as it may be, requires separate optimization for each input image. Our method, on the other hand, allows for a fast prediction via a simple forward-pass operation over the trained network. To demonstrate the efficiency of our approach, we provide a runtime comparison with the analytic solver for each application. We implement the analytic solver in fully-vectorized Numpy-Scipy code, and compare to the network solution, implemented in TensorFlow. Note that in seeded segmentation we compare the overall runtime of calculating the solution of the $128 \times 128$ images, along with the interpolation of the segmentation result back to original size. In other applications we compute the solutions for the original sizes of the test images.
The last row of Table \ref{tbl:results} shows the average evaluation time for each test set using the analytic and network solutions. One can clearly see the benefit in terms of speed in favor of \emph{Deep Energy}, reflected in a speedup factor of $22-56$ across all applications. We should note that our approach enables incorporating more complex energy functions into real-time applications, while avoiding the computationally-heavy direct minimization algorithms.  

\begin{figure*}
    \begin{center}
    \setlength{\tabcolsep}{-1pt}
    \begin{tabular}{cccccc}
    & & $\beta=1000$ & & & \\
    \picsone{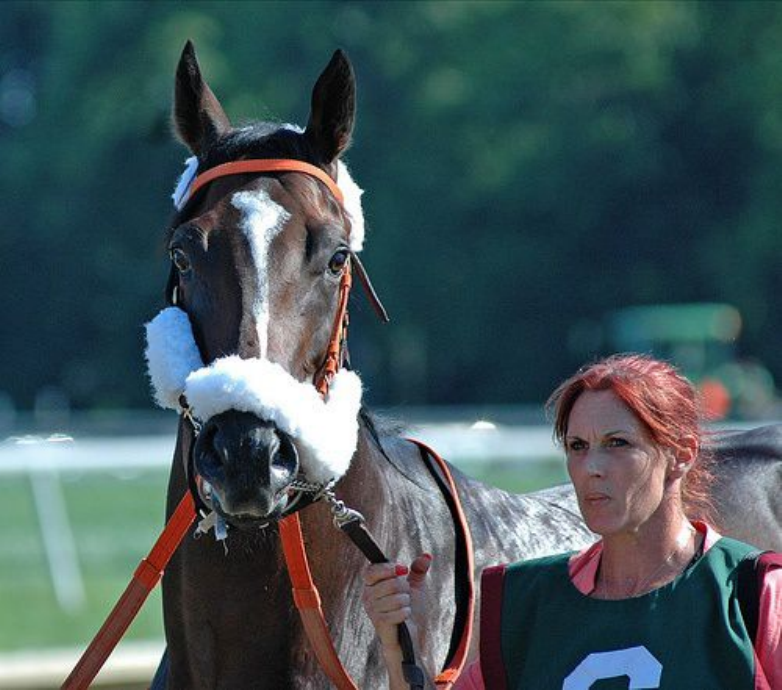} & \picsone{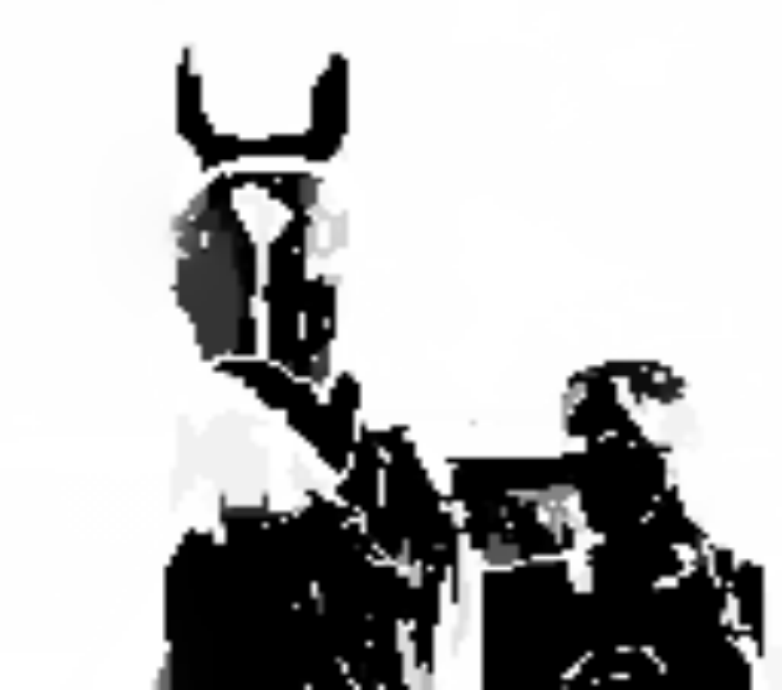} & \picsone{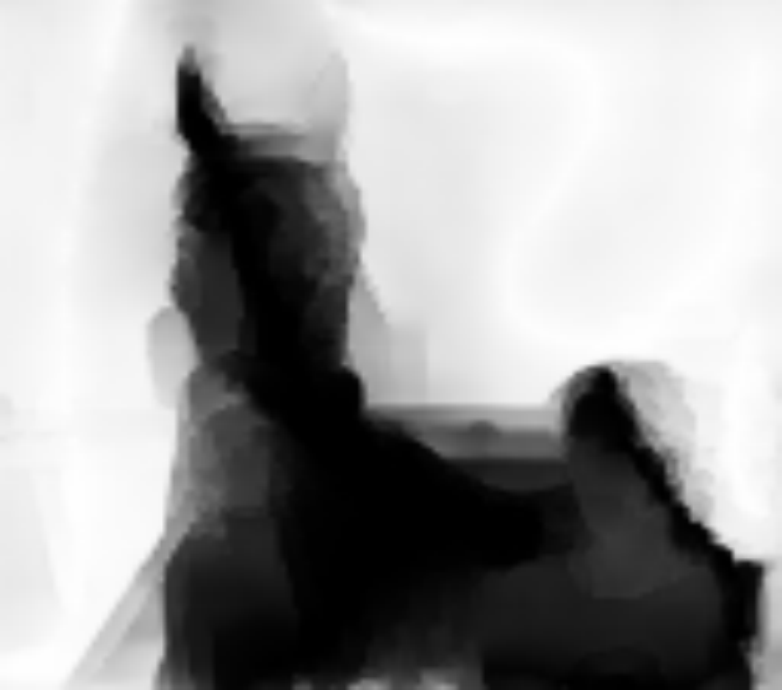} & \picsone{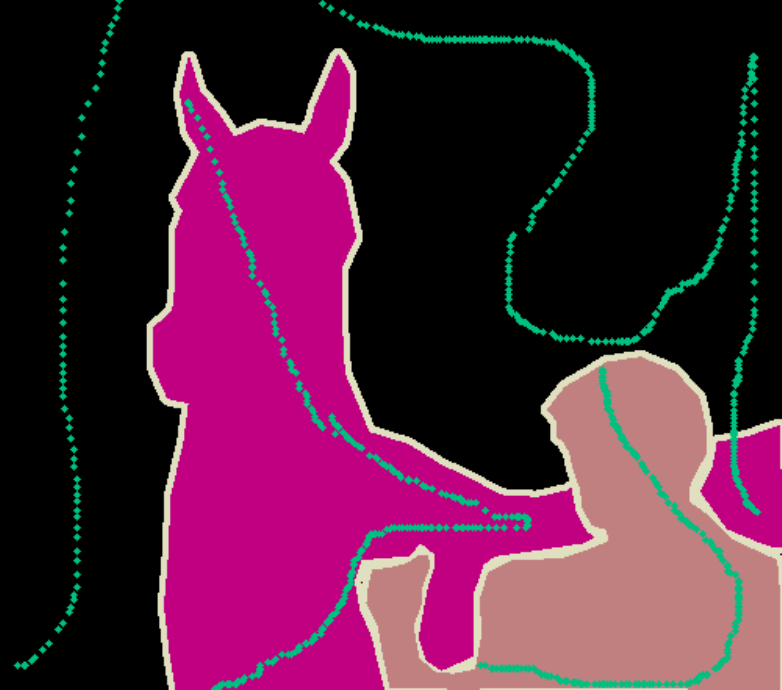} & \picsone{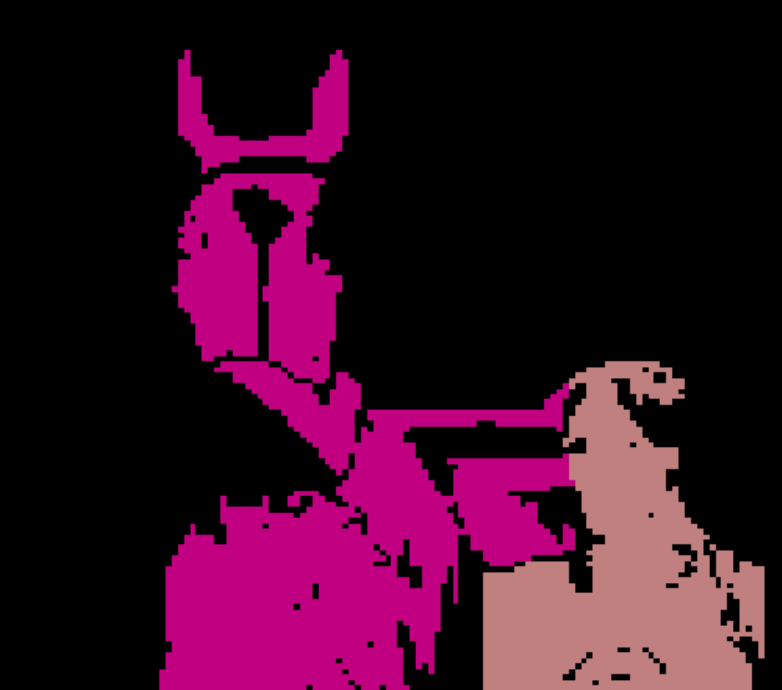} & \picsone{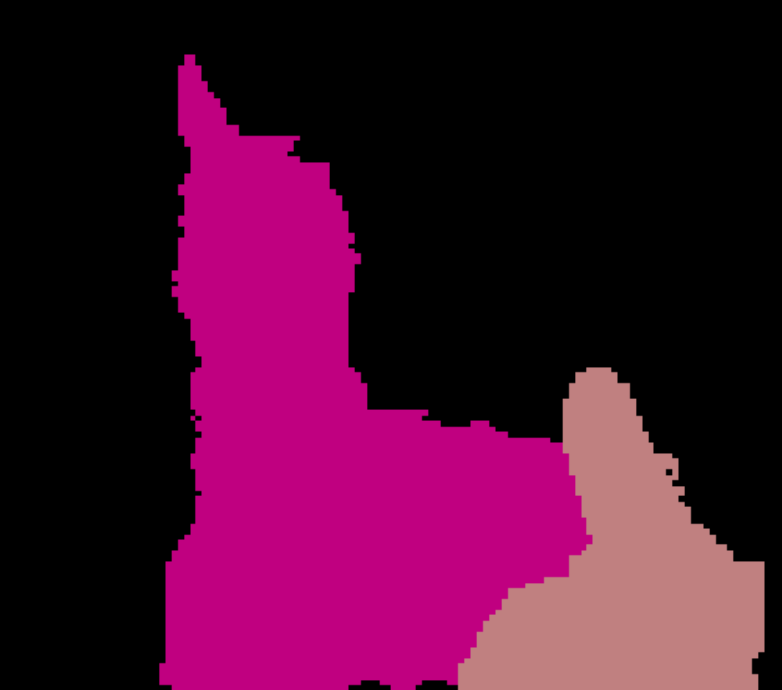} \\ \vspace{-1mm}
    \picsone{4_1} & \picsone{4_2} & \picsone{4_3} & \picsone{4_4} & \picsone{4_5} & \picsone{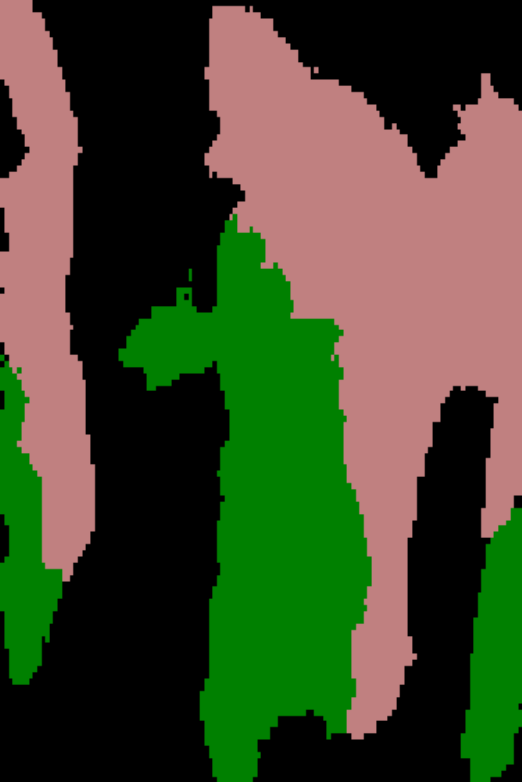} \\ \vspace{-1mm}
    \picsone{16_1} & \picsone{16_2} & \picsone{16_3} & \picsone{16_4} & \picsone{16_5} & \picsone{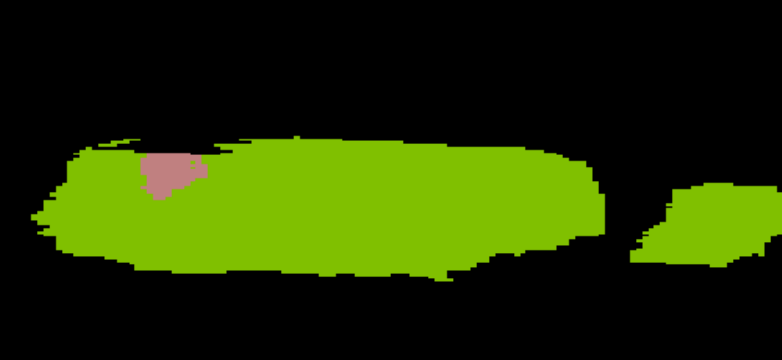} \\ \vspace{-1mm}
    \picsone{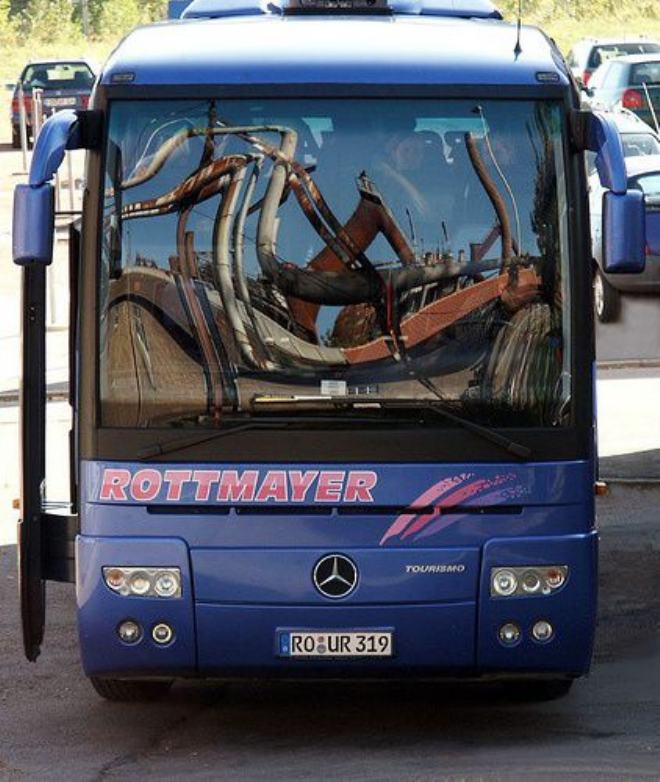} & \picsone{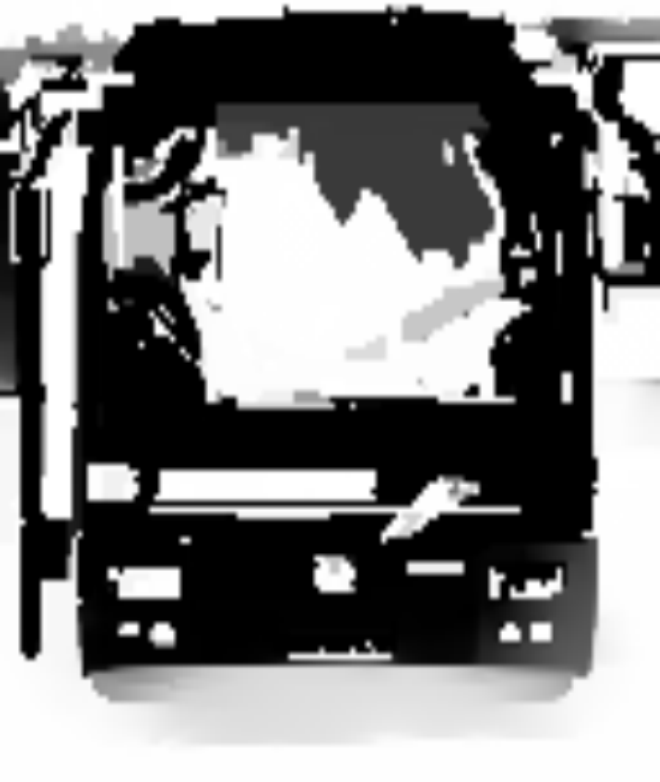} & \picsone{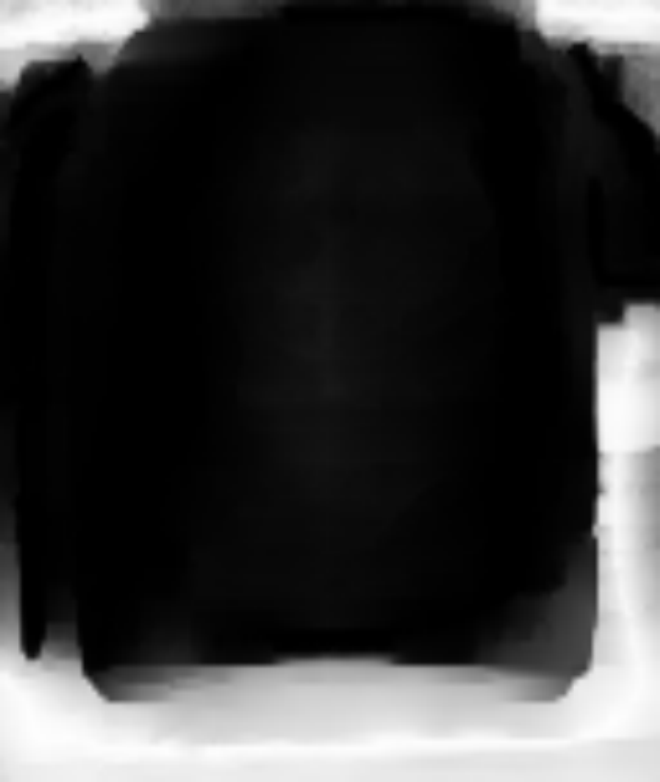} & \picsone{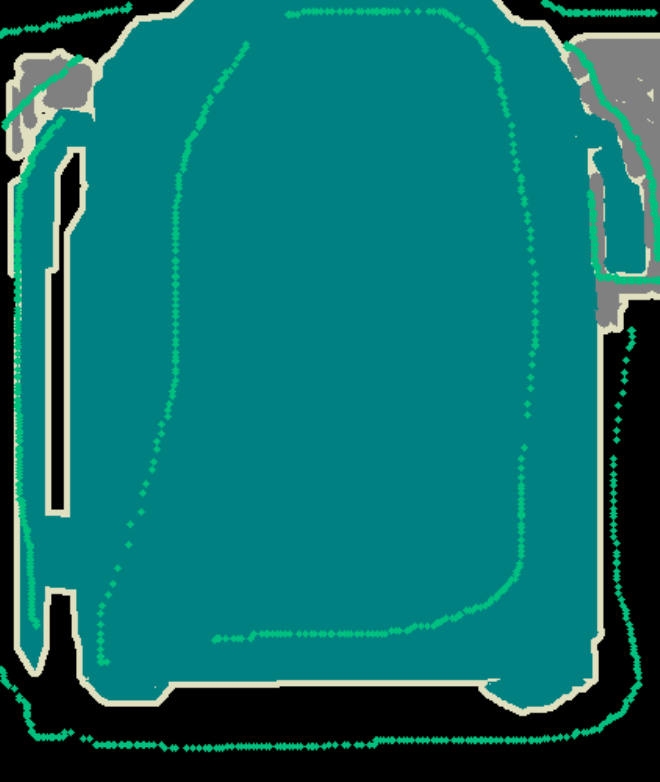} & \picsone{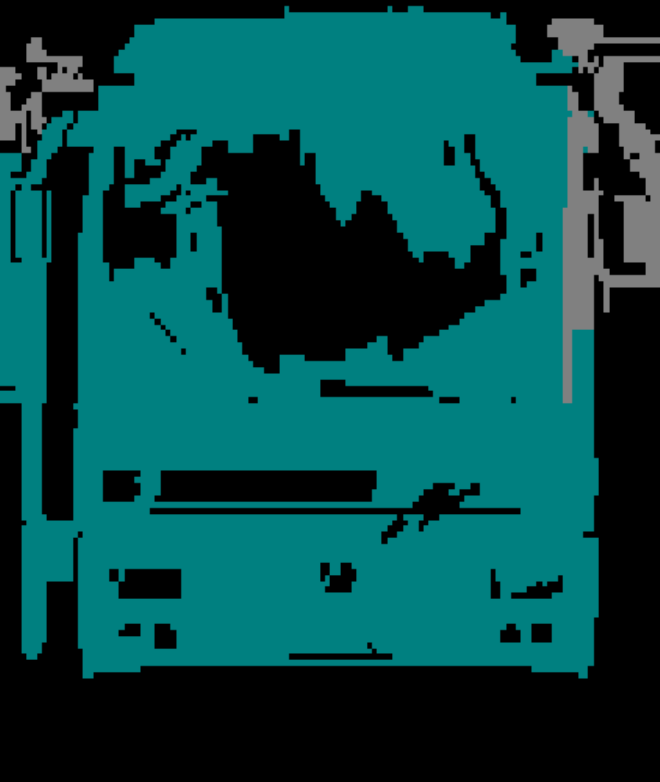} & \picsone{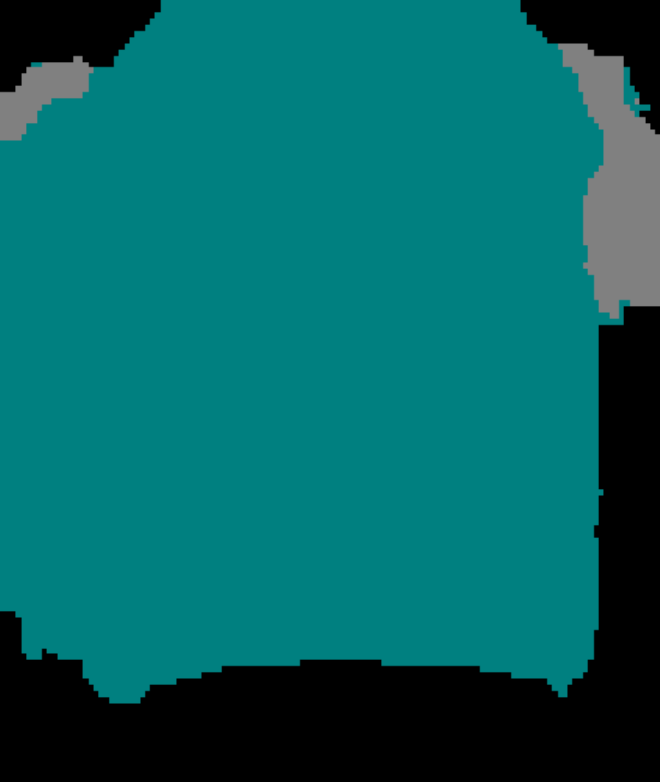} \\ \vspace{-1mm}
    \picsone{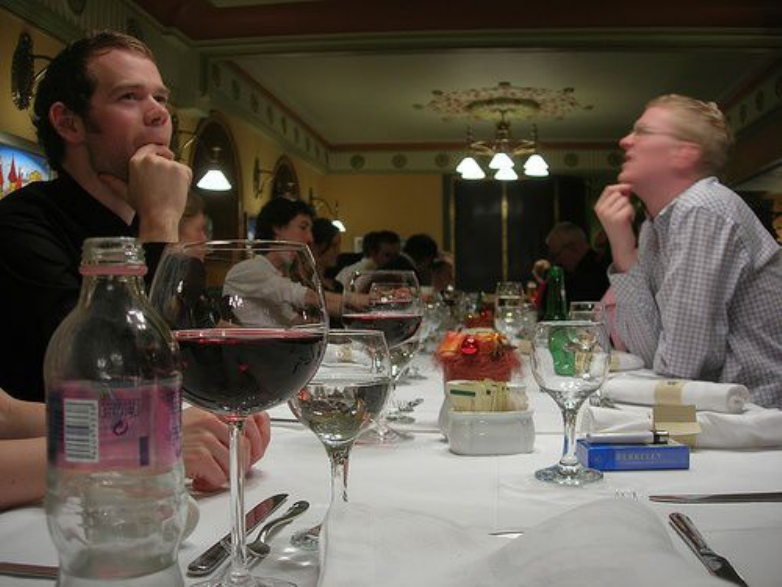} & \picsone{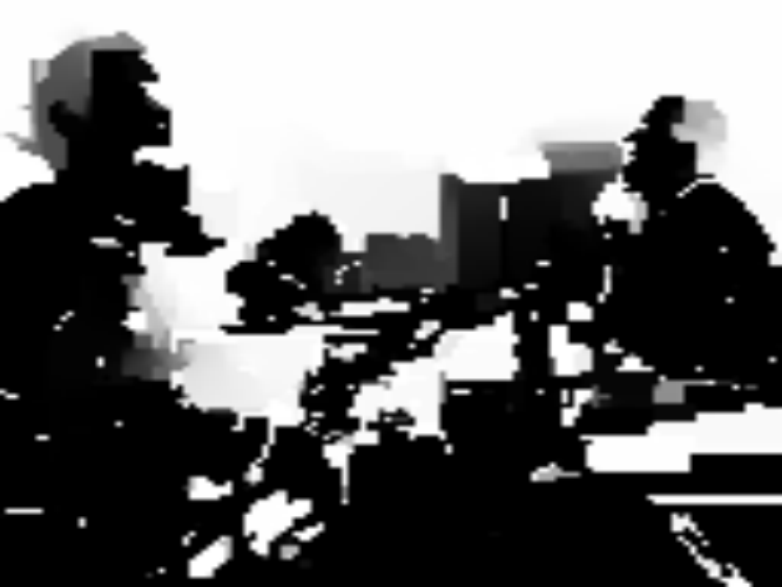} & \picsone{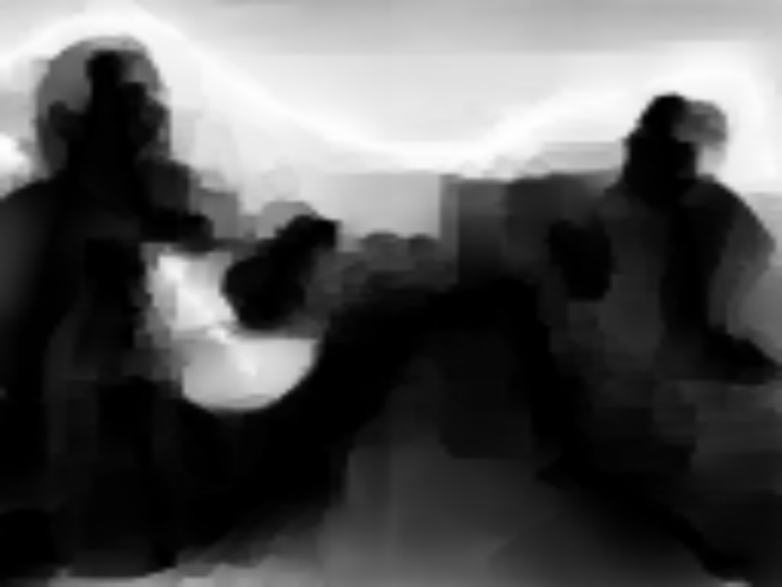} & \picsone{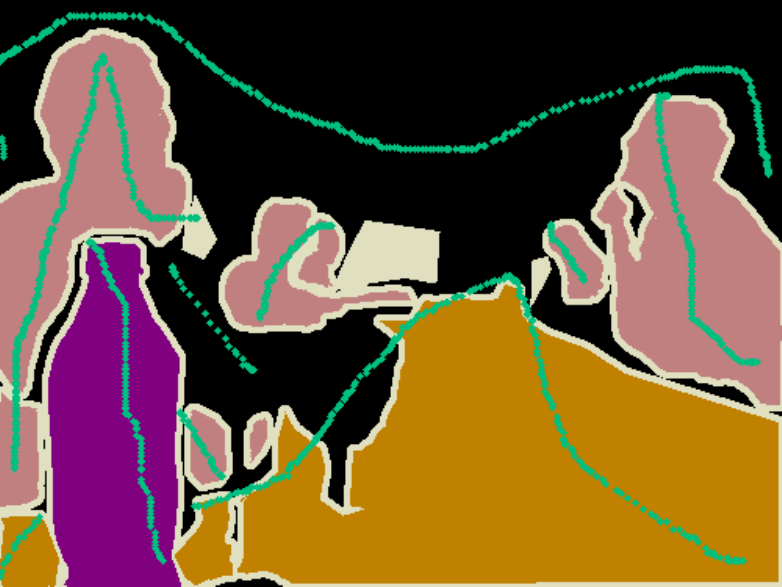} & \picsone{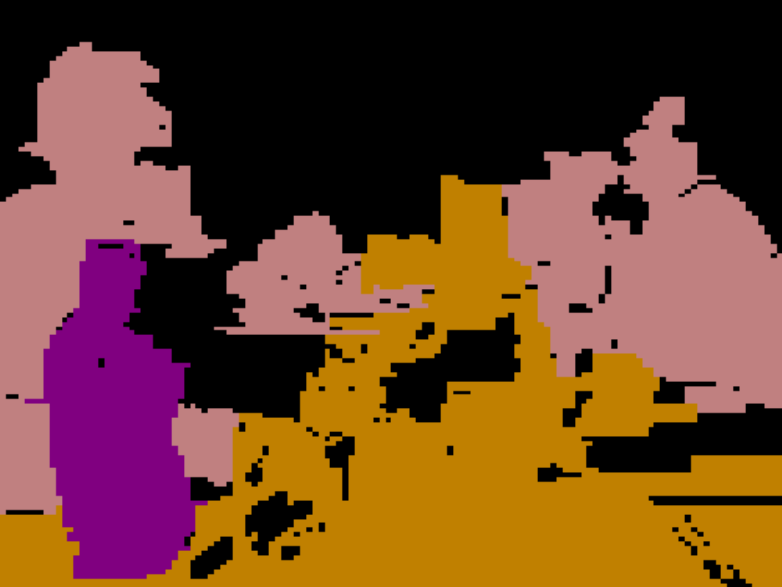} & \picsone{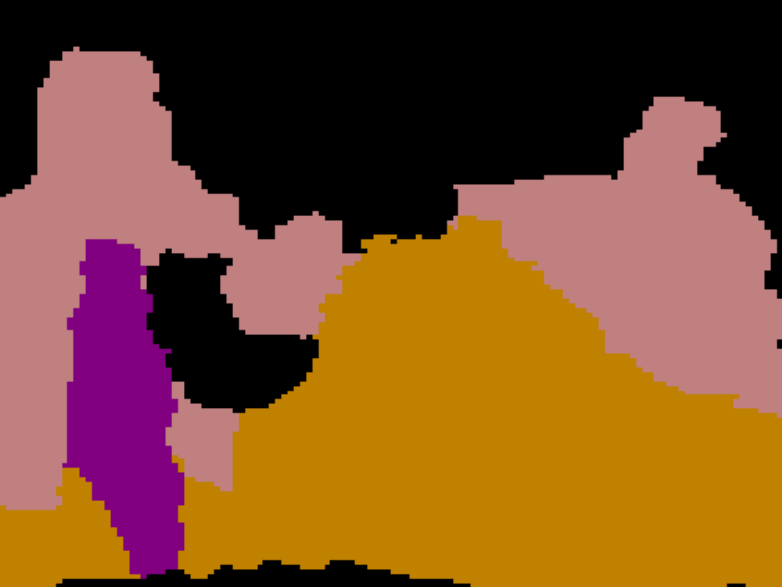} \\ 
    & & $\beta=100$ & & & \\
    \picstwo{1_1} & \picstwo{1_2} & \picstwo{1_3} & \picstwo{1_4} & \picstwo{1_5} & \picstwo{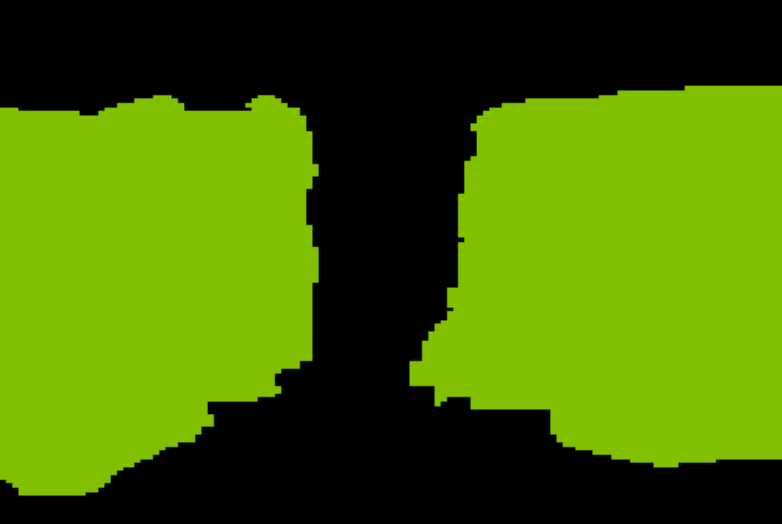} \\ \vspace{-1mm}        			\picstwo{9_1} & \picstwo{9_2} & \picstwo{9_3} & \picstwo{9_4} & \picstwo{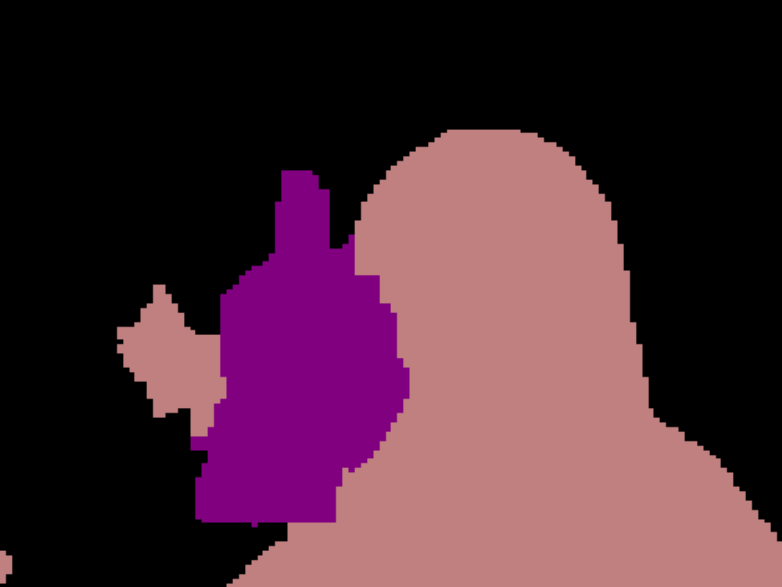} & \picstwo{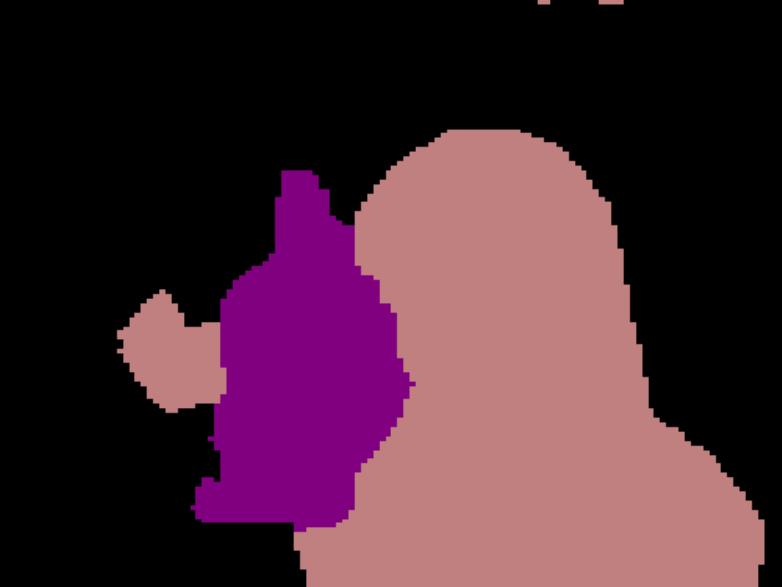} \\\vspace{-1mm}
    \picstwo{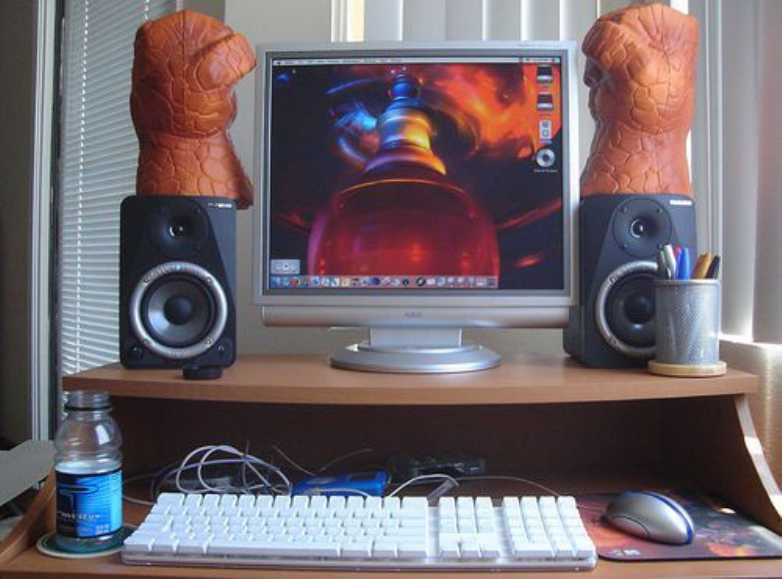} & \picstwo{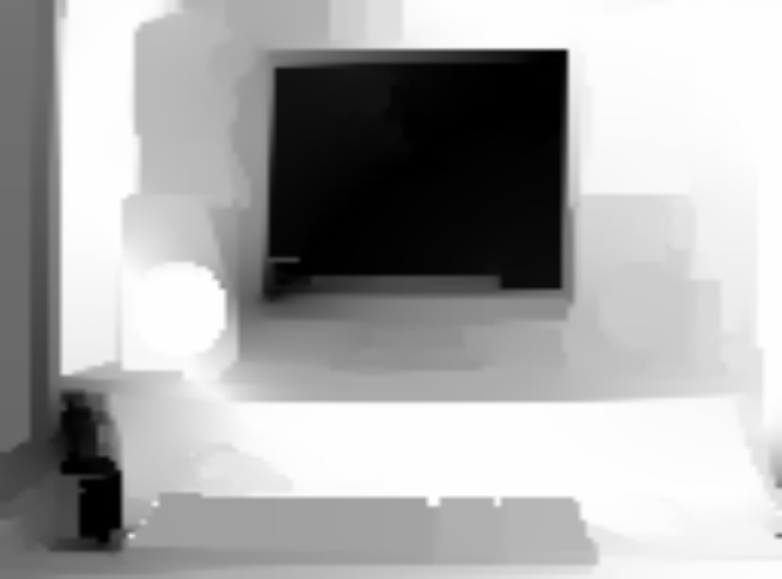} & \picstwo{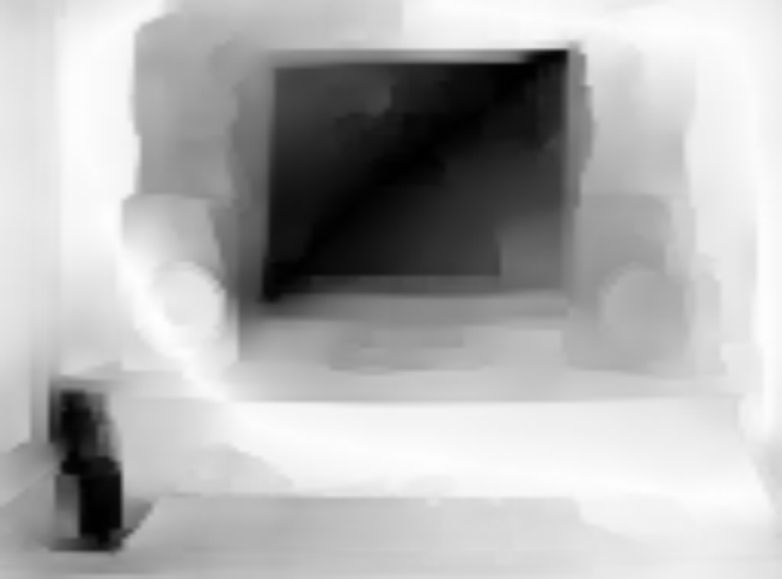} & \picstwo{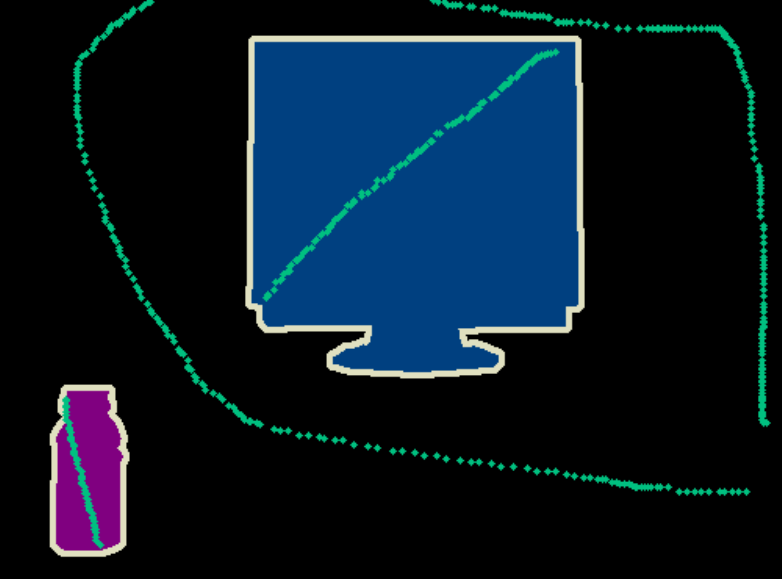} & \picstwo{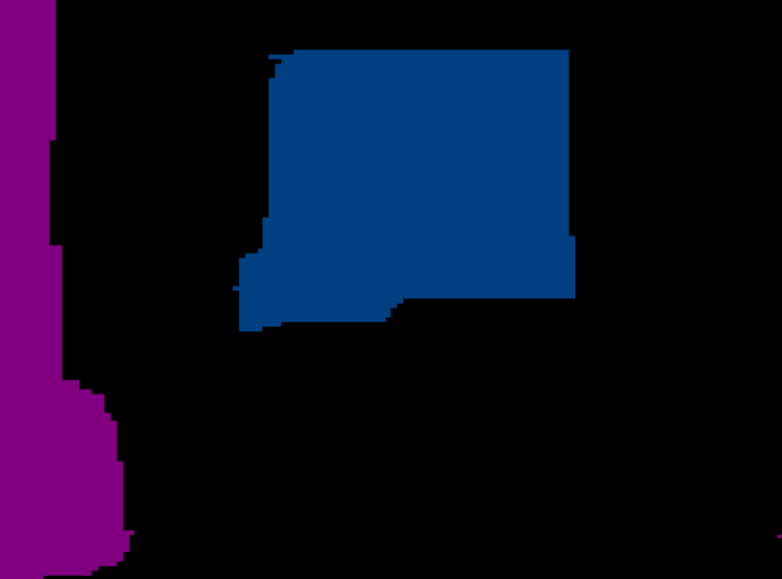} & \picstwo{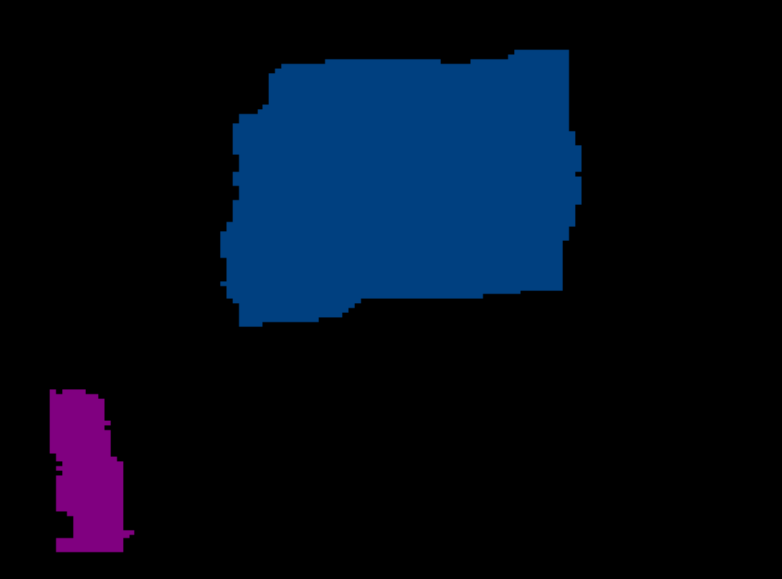} \\\vspace{-1mm}
    \picstwo{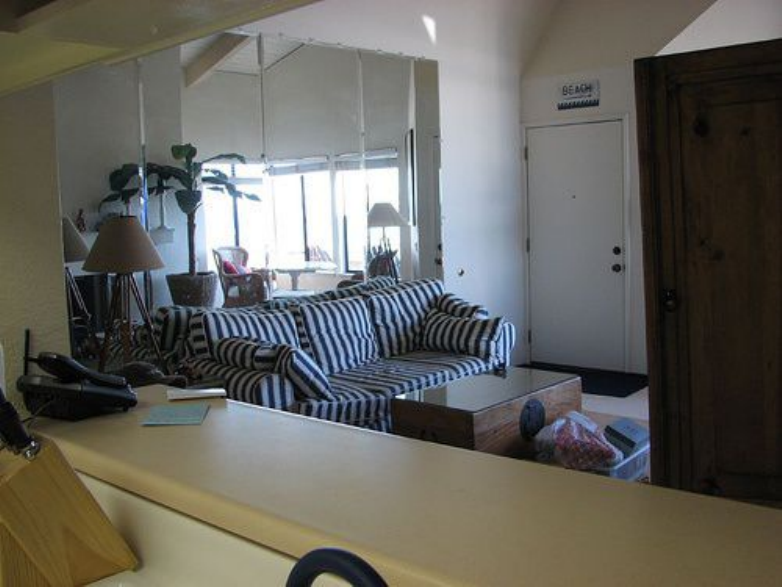} & \picstwo{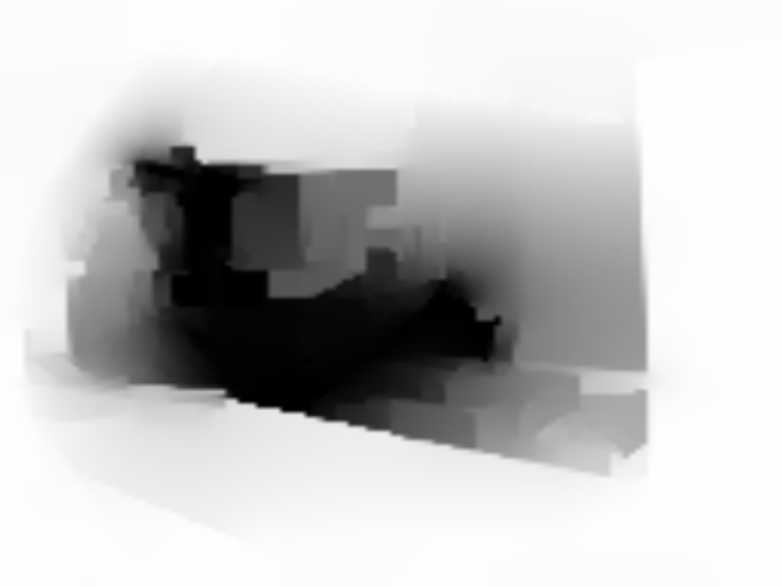} & \picstwo{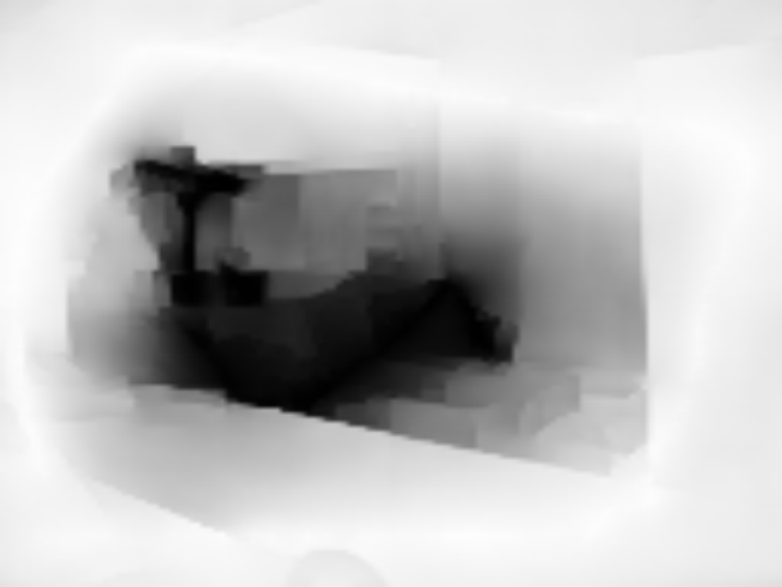} & \picstwo{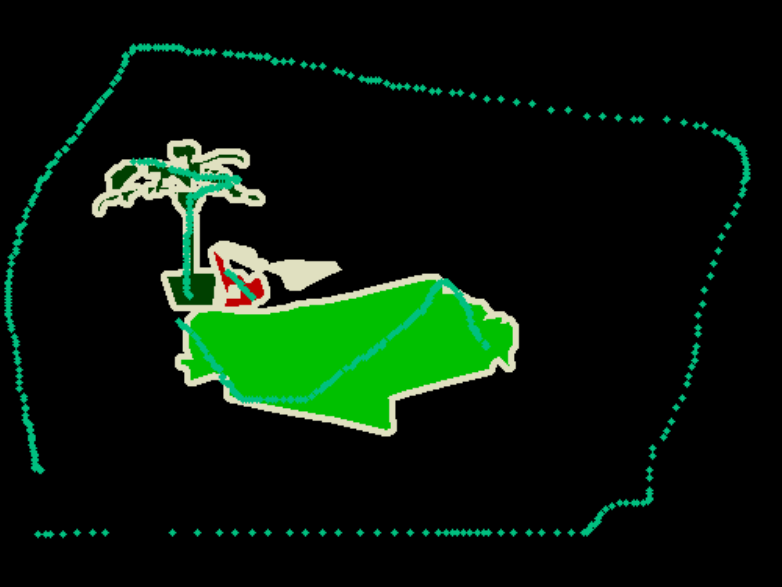} & \picstwo{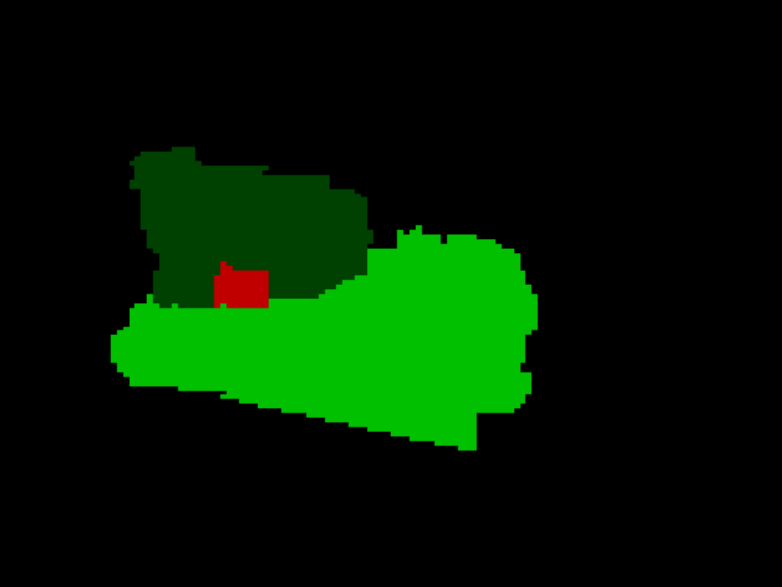} & \picstwo{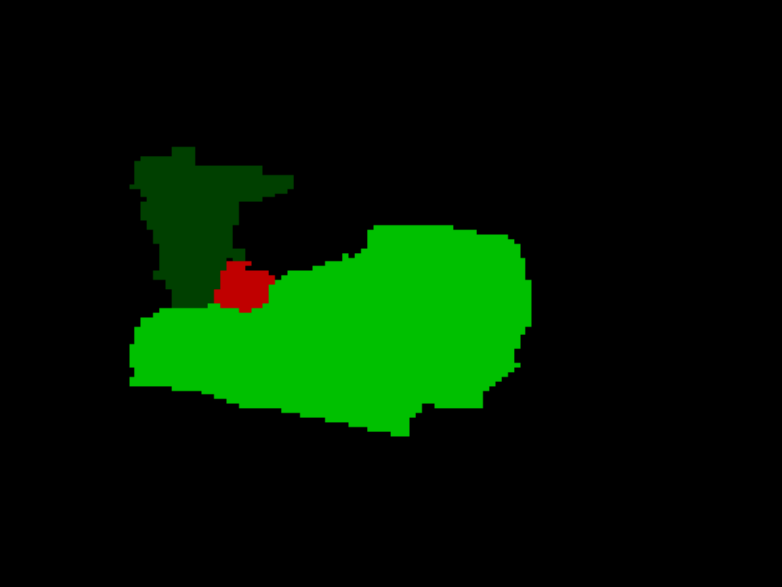} \\\vspace{-1mm}
    \picstwo{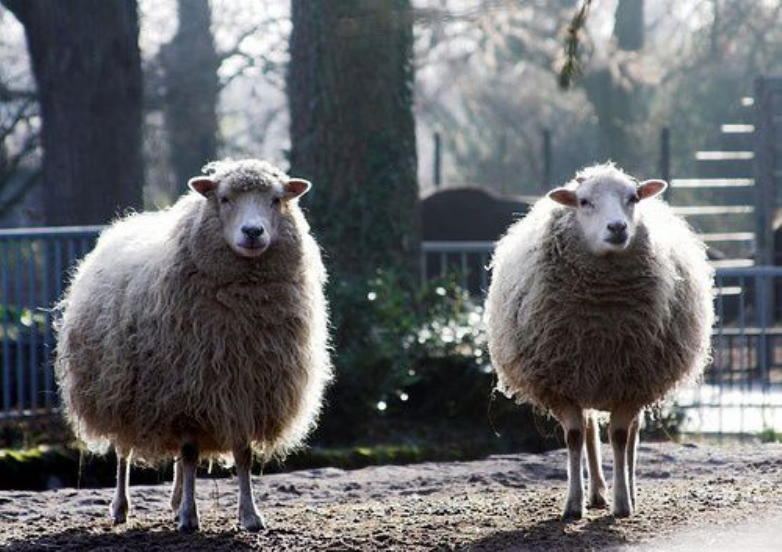} & \picstwo{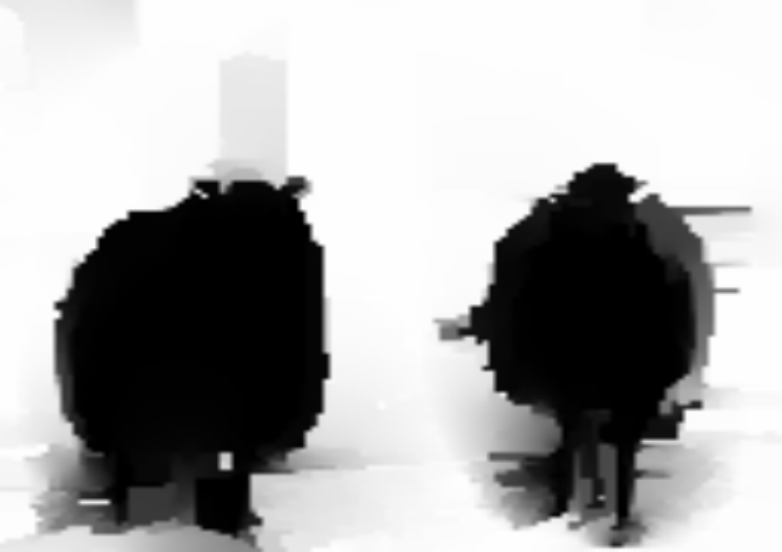} & \picstwo{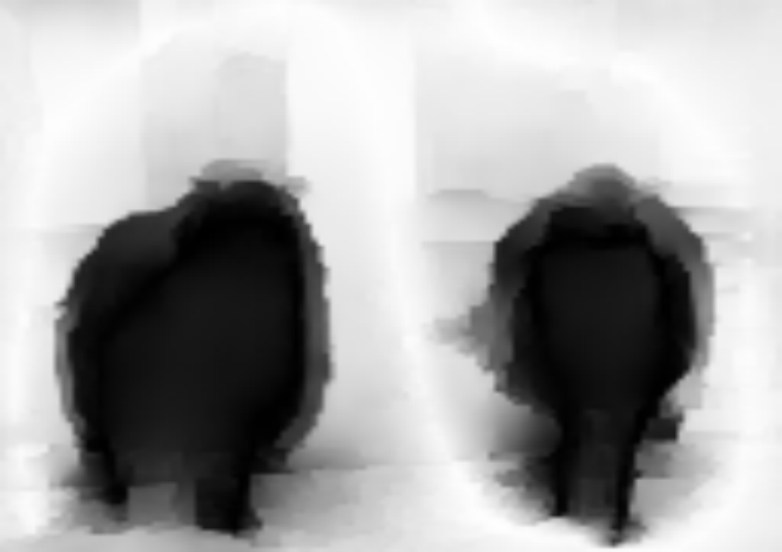} & \picstwo{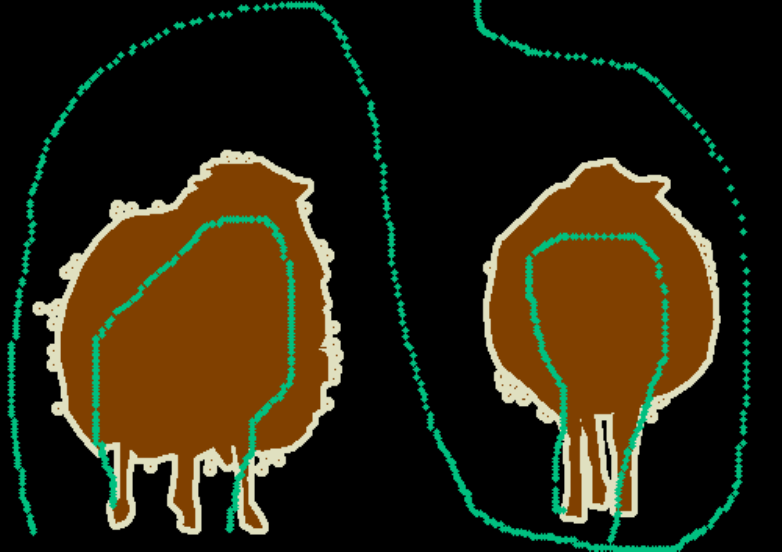} & \picstwo{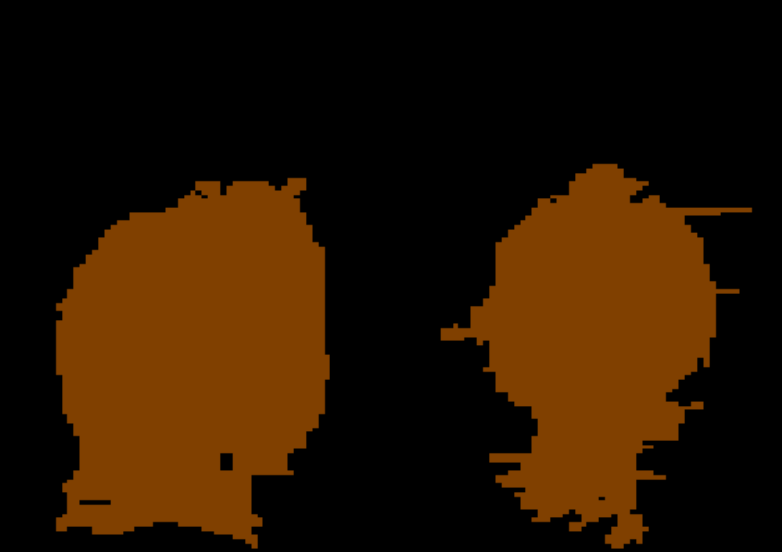} & \picstwo{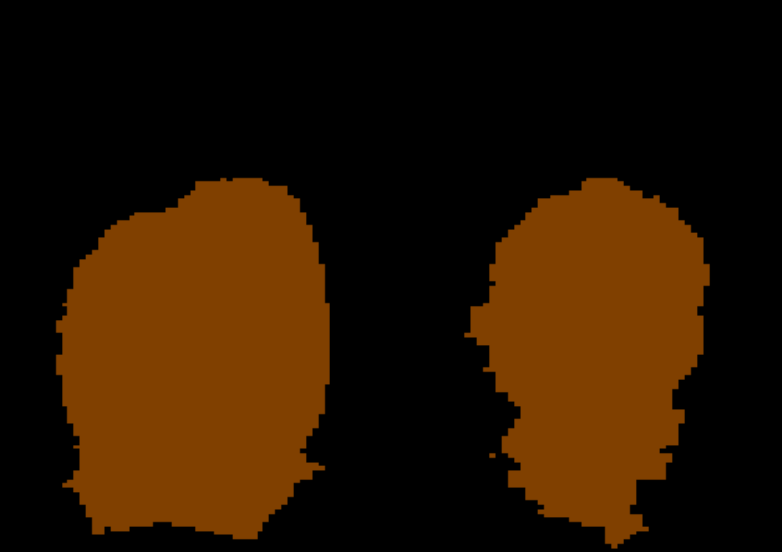} \\\vspace{-1mm}
    \small{input} & \small{analytic prob.} & \small{network prob.} & \small{GT + seeds} & \small{analytic seg.} & \small{network seg.} \\
    \end{tabular}
    \end{center}
    \caption{Seeded segmentation results with different $\beta$ values. From left to right: original image, analytic and network probabilities of `background' class, ground truth segmentation with overlayed seeds, analytic and network segmentation.}
    \label{fig:segmentation_results}
\end{figure*}

\begin{figure}
    \begin{center}
    \setlength{\tabcolsep}{-1pt}
    \begin{tabular}{ccccc}
    \vspace{-1mm}
    \picm{6_1} & \picm{6_2} & \picm{6_3} & \picm{6_4} & \picm{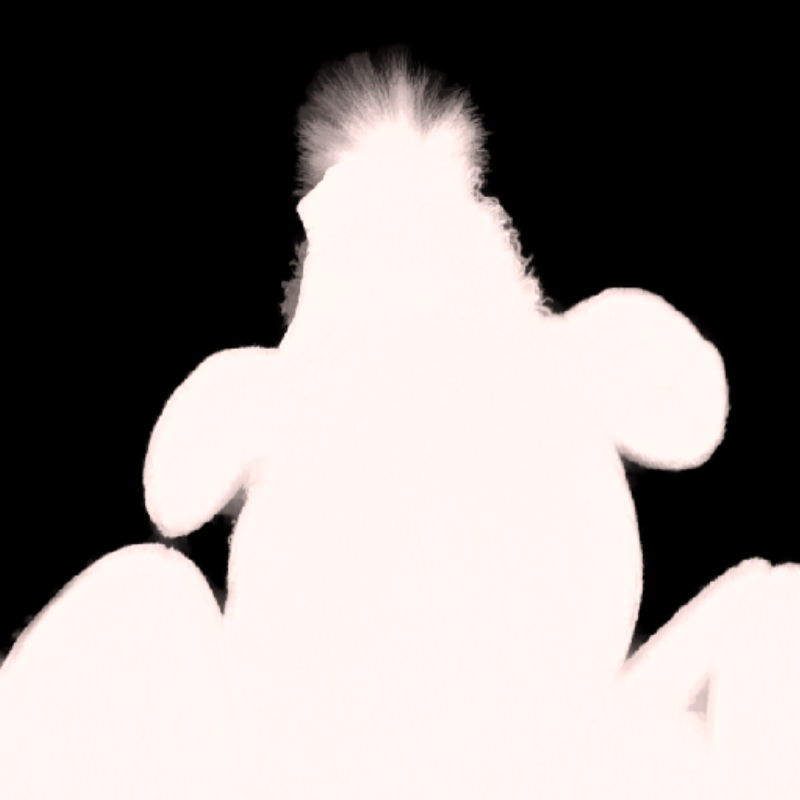} \\ \vspace{-1mm}
    \picm{16_1} & \picm{16_2} & \picm{16_3} & \picm{16_4} & \picm{16_5} \\ \vspace{-1mm}
    \picm{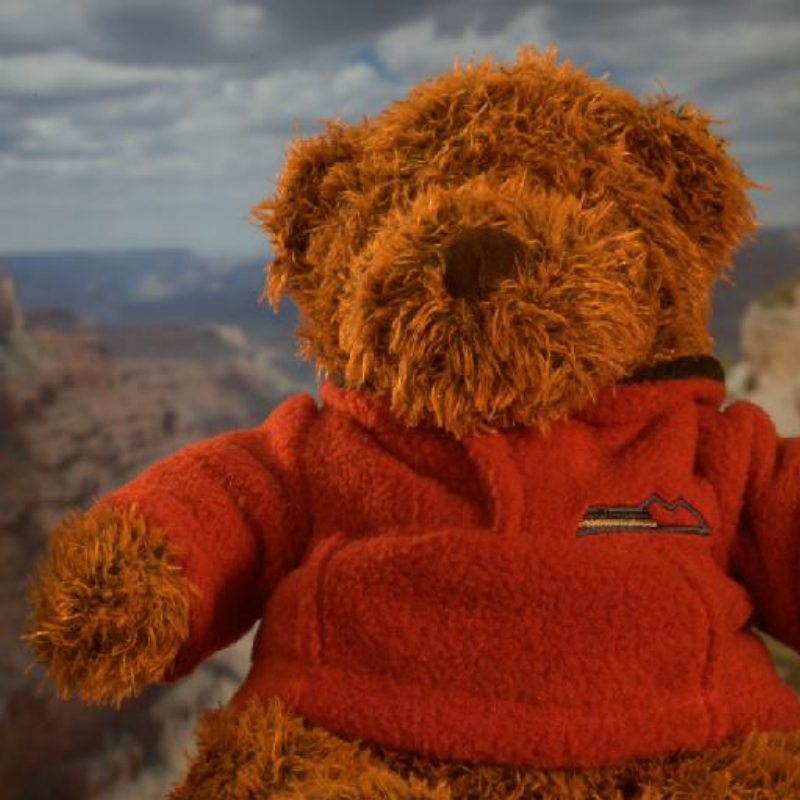} & \picm{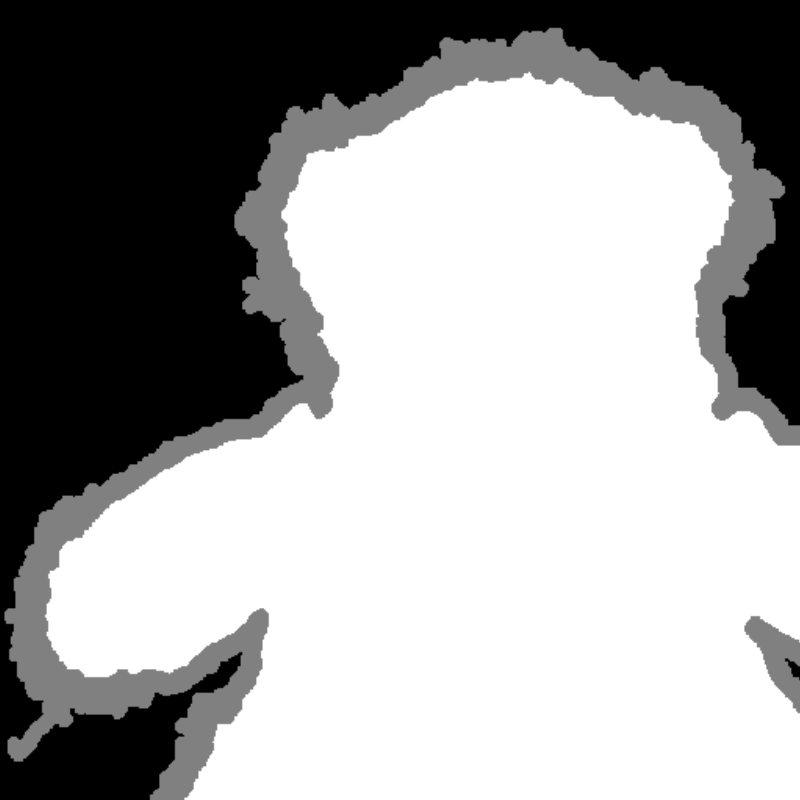} & \picm{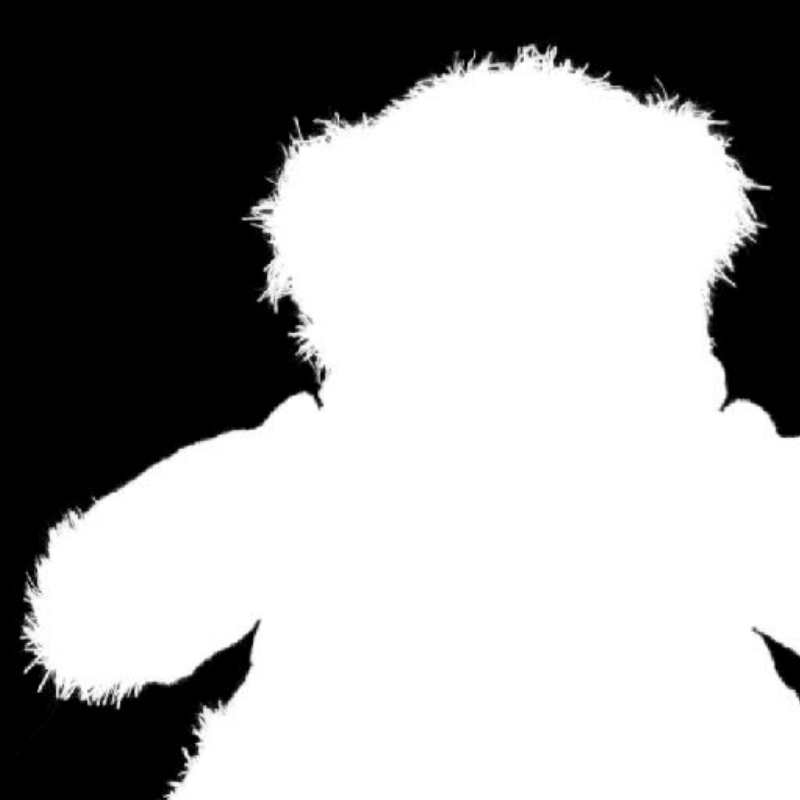} & \picm{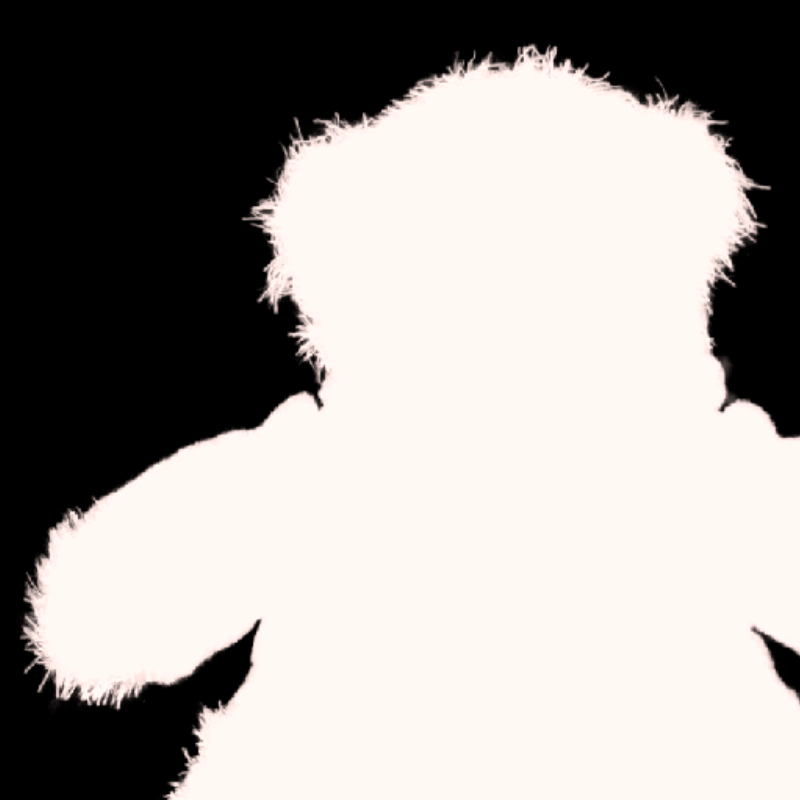} & \picm{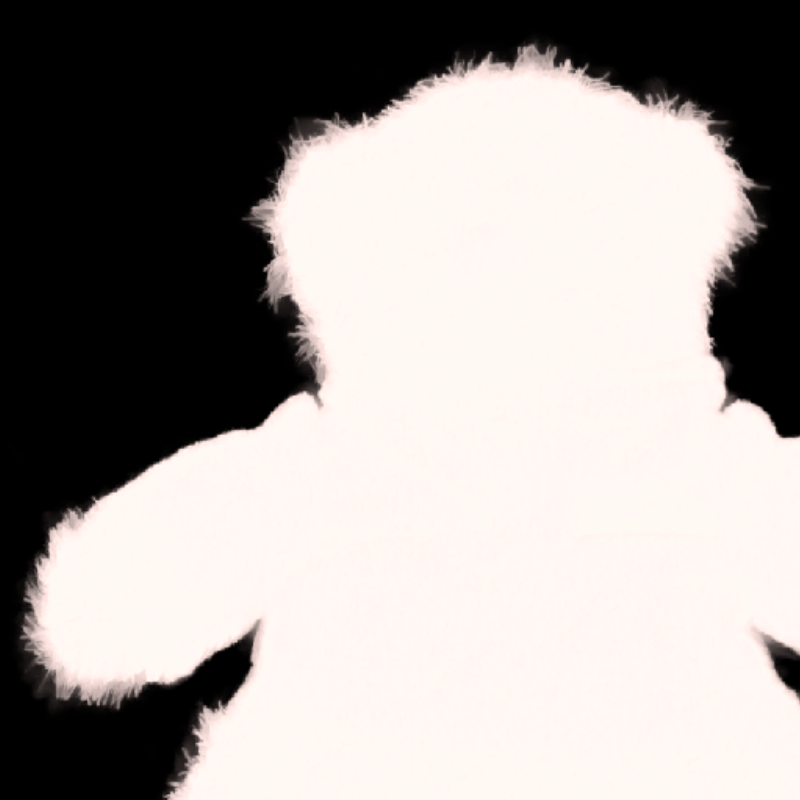} \\ \vspace{-1mm}
    \picm{4_1} & \picm{4_2} & \picm{4_3} & \picm{4_4} & \picm{4_5} \\ \vspace{-1mm}
    \picm{1_1} & \picm{1_2} & \picm{1_3} & \picm{1_4} & \picm{1_5} \\ \vspace{-1mm}
    \picm{3_1} & \picm{3_2} & \picm{3_3} & \picm{3_4} & \picm{3_5} \\ \vspace{-1mm}
    input & trimap & GT & analytic & network
    \end{tabular}
    \end{center}
    \caption{Image matting results. From left to right: original image, input trimap, matting ground truth, analytic solution and network solution.}
    \label{fig:matting_results}
\end{figure}

\begin{figure}
    \begin{center}
    \setlength{\tabcolsep}{-1pt}
    \begin{tabular}{cccc}
    \vspace{-1mm}
    \picd{1_4} & \picd{1_2} & \picd{1_3} & \picd{1_1} \\ \vspace{-1mm}
    \picd{2_4} & \picd{2_2} & \picd{2_3} & \picd{2_1} \\ \vspace{-1mm}
    \picd{3_4} & \picd{3_2} & \picd{3_3} & \picd{3_1} \\ \vspace{-1mm}
    \picd{4_4} & \picd{4_2} & \picd{4_3} & \picd{4_1} \\ \vspace{-1mm}
    \picd{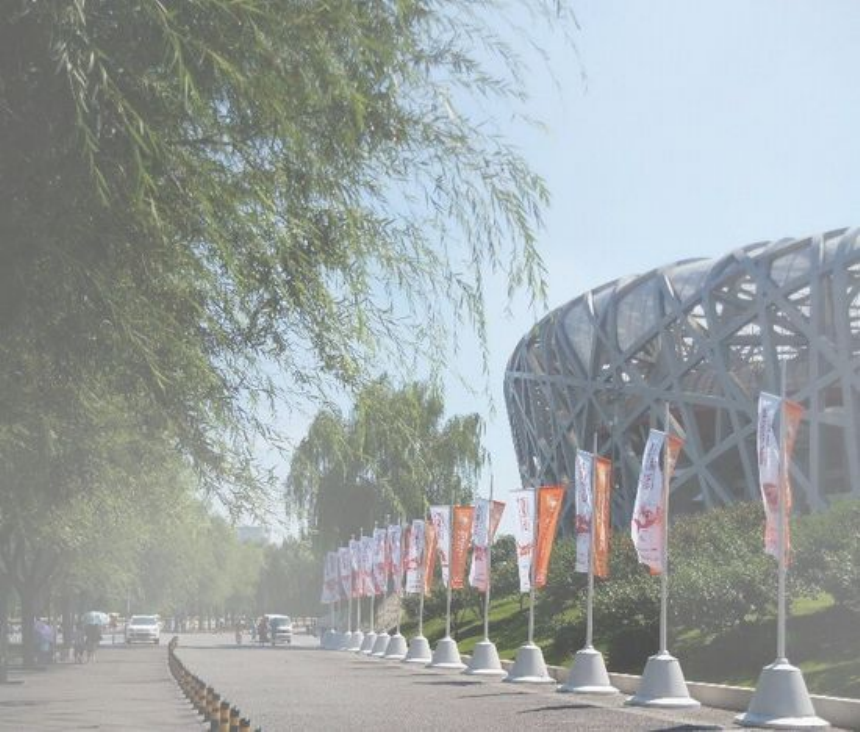} & \picd{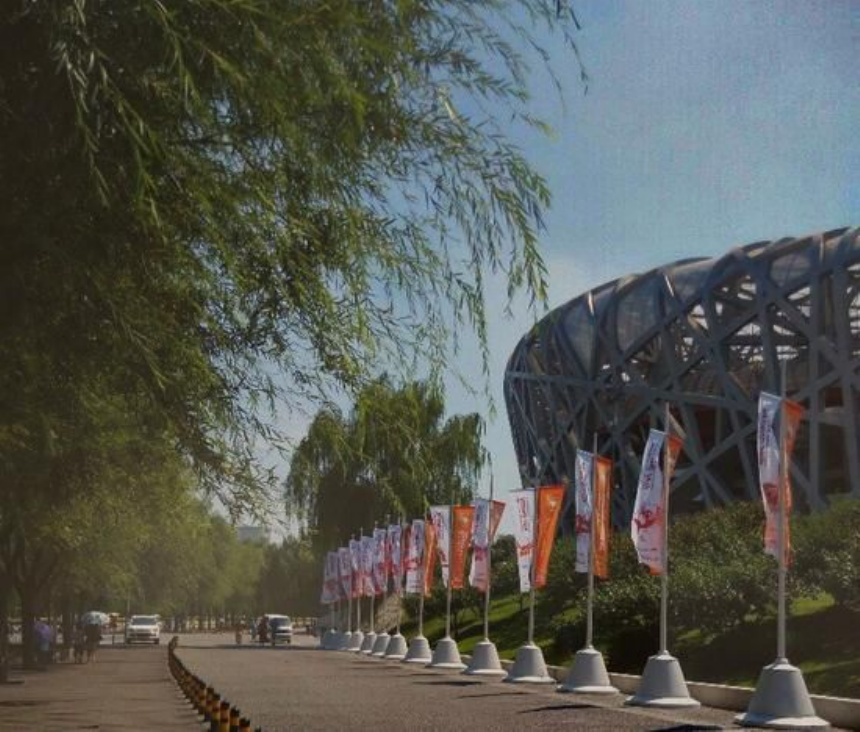} & \picd{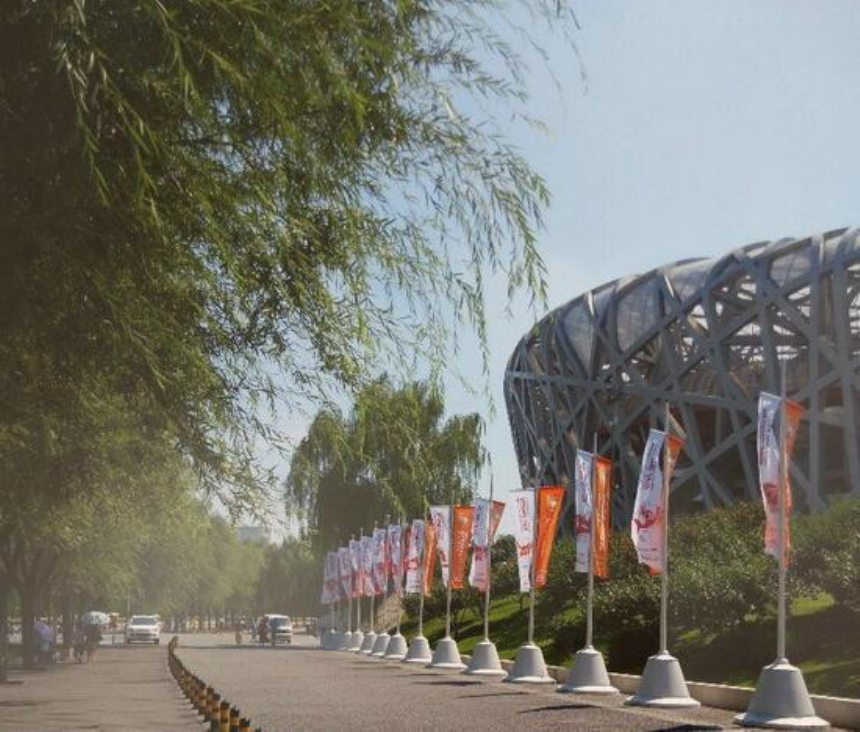} & \picd{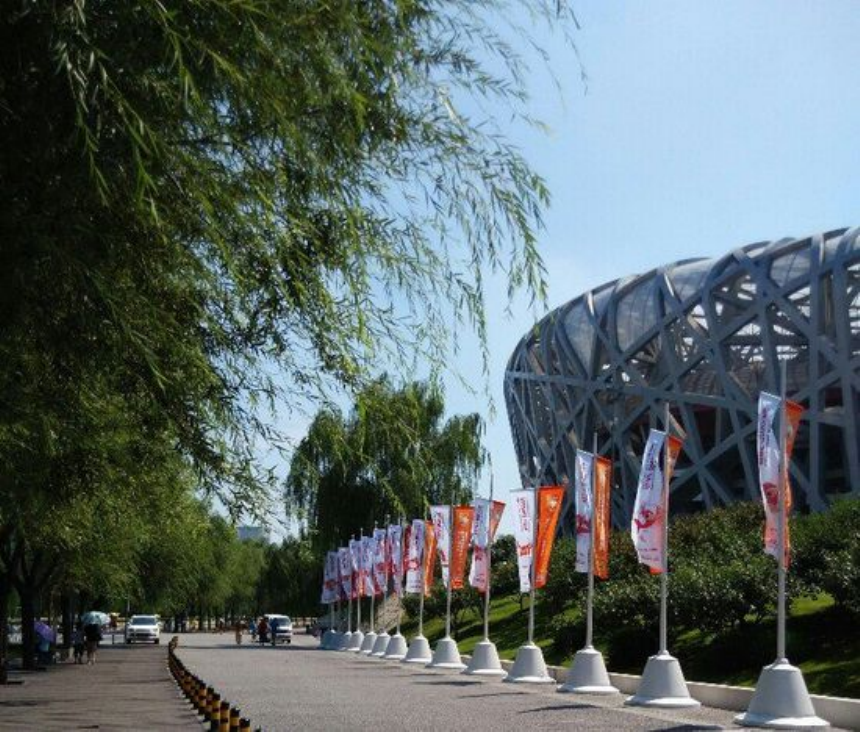} \\ \vspace{-1mm}
    input & analytic & network & GT
    \end{tabular}
    \end{center}
    \caption{Single image dehazing results. From left to right: hazy image, analytic solution, network solution and ground truth. }
    \label{fig:dehazing_results}
\end{figure}

\begin{figure}
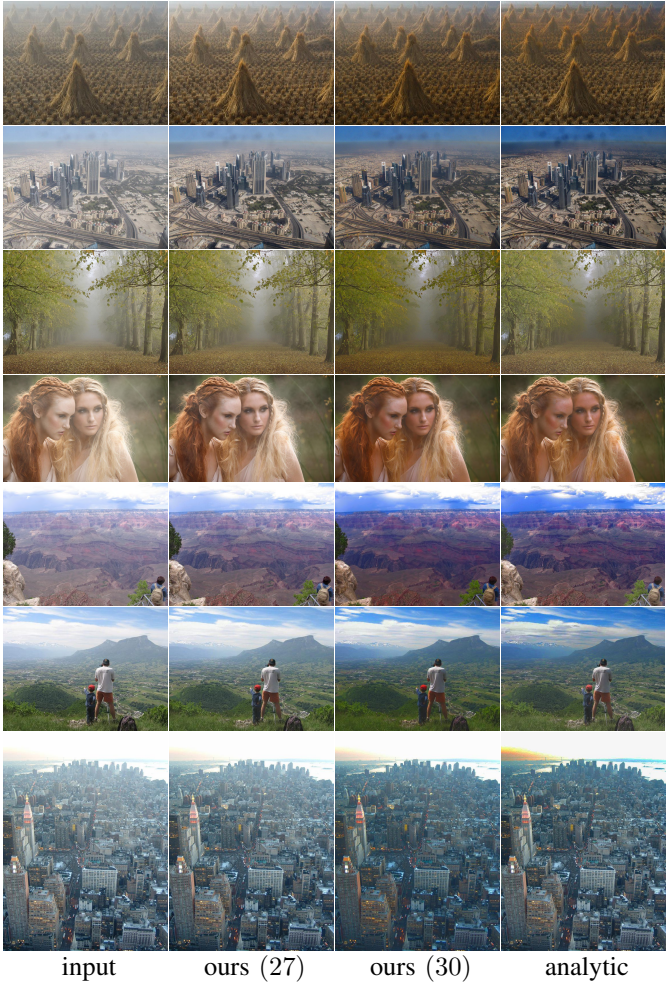

    \begin{center}
    \setlength{\tabcolsep}{-1pt}
    \begin{tabular}{cccc}
    \vspace{-1mm}
    \picdr{1_1} & \picdr{1_3} & \picdr{1_4} & \picdr{1_2} \\ \vspace{-1mm}
    \picdr{2_1} & \picdr{2_3} & \picdr{2_4} & \picdr{2_2} \\ \vspace{-1mm}
    \picdr{3_1} & \picdr{3_3} & \picdr{3_4} & \picdr{3_2} \\ \vspace{-1mm}
    \picdr{4_1} & \picdr{4_3} & \picdr{4_4} & \picdr{4_2} \\ \vspace{-1mm}
    \picdr{6_1} & \picdr{6_3} & \picdr{6_4} & \picdr{6_2} \\ \vspace{-1mm}
    \picdr{8_1} & \picdr{8_3} & \picdr{8_4} & \picdr{8_2} \\ \vspace{-1mm}
    \picdr{9_1} & \picdr{9_3} & \picdr{9_4} & \picdr{9_2} \\ \vspace{-1mm}
    input & ours $(27)$ & ours $(30)$ & analytic
    \end{tabular}
    \end{center}
    \caption{Single image dehazing results on real-world images. From left to right: hazy image, analytic solution, network solution after $27$ epochs and after $30$ epochs.}
    \label{fig:dehazing_real_results}
\end{figure}


\subsection{Qualitative Results} \label{ss:qualitative}

We turn to present the results of seeded segmentation on images taken from Pascal VOC 2012 `val' dataset, accompanied with seeds collected by \cite{scribblesup}. The upper and lower parts of Figure\footnote{Here and elsewhere, visual results are best viewed with zoom of $300\%$.} \ref{fig:segmentation_results} feature results with $\beta=1000$ and $\beta=100$ respectively. For the analytic solution, higher values of $\beta$ lead to better localized, but often unstable, results (e.g., holes within objects and over-segmentation), thus a lower value of $\beta=100$ is optimal. The network solution, however, returns blurrier probability maps, thus an initially higher value of $\beta=1000$ creates a better balance. Overall, the analytic solution, by its construction, is highly affected by local grayscale changes in the image. For example, it is strongly affected by the white straps around the horse's head in the first row, or the train window in the sixth row. The network solution, on the other hand, seems to better perform in these cases, due to the extra regularization induced by the chosen network. While the initial energy function used for the analytic and network solutions are the same, the solutions may differ and depend on the hyper-parameters and the network's architecture.

In image matting, we show the results of images from the training set of \textit{alphamatting.com} \cite{alpha_matting_com}, which feature heavier user assistance in the form of trimaps. Although we trained on Pascal VOC 2012, originally intended for image segmentation, and used minimal seeds, our network solution, shown in Figure \ref{fig:matting_results}, exhibits good generalization ability. In the top four rows, the analytic and network solutions appear similar, although the network returns slightly blurrier results. In the fifth row, the holes of the decorative object are completely filled by the analytic solution, whereas the network solution is closer to the ground truth. The last row showcases the failure of the network in capturing the very fine hair on the trolls' heads. This blurriness is a result of the architecture that features a long chain of repeated convolutions. It can possibly be remedied by an additional ``refinement network'' as suggested in \cite{deep_matting} and adapted to the \emph{Deep Energy} loss. We leave this important direction for future work. 

We present the qualitative performance for the single image dehazing problem on RESIDE's HSTS benchmark images. The analytic solution amplifies the contrast and colors in the image, especially in the sky regions, where the dark channel prior assumption is not valid. Our solution, on the other hand, returns a more realistic result, much similar to the ground truth. While we train our network using the DCP energy function, our aim is not reaching the absolute global minimum of the energy function. By an early-stopping during training, we achieve an additional  effective regularization and provide even better results than DCP. 

Figure \ref{fig:dehazing_real_results} shows real-world image results of our network, trained for $27$ and $30$ epochs. As can be seen, as training progresses the network reaches a lower minimum of the energy function, providing similar results as classical DCP. Early stopping after $27$ epochs provides a more subtle dehazing effect without an enhanced contrast, but with residual haze. By changing the training stopping point, we can control the proximity of the network to the analytic solution of DCP.


\section{Conclusion} \label{s:conclusion}

This work has introduced a new and principled approach for unsupervised training of DNNs, through direct minimization of energy functions that describe the desired inference. The proposed \emph{Deep Energy} paradigm has been demonstrated on three applications: seeded segmentation, image matting and single image dehazing. Our approach incorporates task-specific domain knowledge into the loss function and allows for reduced dependency on annotated and synthetic datasets. We have demonstrated that even though we train our network to approximate the minimization of a certain energy function, our solution is faster and often of much better quality, compared to that of the analytic alternative. This implies an effective regularization, which may stem either from the architecture itself or from the learning process. Future directions of research include combining supervised and energy-based losses, investigating the source of added regularization, and a derivation of a theoretical comparison between supervised and energy-based training.


\appendix[Tensorization of Energy Functions] \label{s:appendix}
\smallpar{Seeded Segmentation} One can directly use the energy expression in Equation (\ref{eq:Energy}) as a loss function during training. Instead, we convert this function to a more ``tensor-friendly'' form using a common expansion of Laplacian matrices:
\begin{equation}
\begin{split}
E(\bX,\bY) & = \frac{1}{2}\sum_l \sum_{(i,j) \in \mathcal{E}} w_{ij} (y_i^{l}-y_j^{l})^2 \\
           & +\lambda \sum_i \left(\sum_l x_i^{l}\right)\left(\sum_l\left(y_i^{l}-x_i^{l}\right)^2\right). \\
\end{split}
\end{equation}
Then, we concatenate $C$ identical copies of the output $\bY \in \mathbb{R}^{N \times L}$, along the last dimension, to create $\tilde{\bY} \in \mathbb{R}^{N \times L \times C}$, where $C$ is the number of neighbors ($C=4$ in our case). Further, we form $\tilde{\bY}_\eta \in \mathbb{R}^{N \times L \times C}$ as a concatenation of the $C$ ``neighbor images'' of the output $\by^l$; in the case of 4-neighborhoods,  the neighbor images are simply the image $\by^l$ shifted left, right, up and down. Finally, the weights can be represented as an $N \times C$ matrix; we then take $L$ copies of this matrix and put them in the 3-tensor $\bW \in R^{N \times L \times C}$. The energy function can now be written as follows:
\begin{equation}\label{eq:segmentation_final}
\begin{split}
E(\bX,\bY) &= \frac{1}{2}\sum_{b=1}^B \sum_{n=1}^N \sum_{l=1}^L \sum_{c=1}^C \bW \odot (\tilde{\bY}-\tilde{\bY}_\eta)^2 \\
           &+\lambda \sum_{b=1}^B \sum_{n=1}^N \left[\sum_{l=1}^L \bX\right] \odot \left[\sum_{l=1}^L(\bY-\bX)^2\right], \\
\end{split}
\end{equation}
where we have summed over the batch dimension, $b=1...B$; and the powers are taken elementwise.

\smallpar{Image Matting} We again rephrase the first term in the energy function in Equation (\ref{eq:matting_energy}) in terms of weights:
\begin{equation}
\balpha ^T \bL \balpha = \sum_{n=1}^N \sum_{i=1}^9 \sum_{j=1}^9 w_{ij}^n (\alpha_i - \alpha_j)^2,
\end{equation}
where we sum over all overlapping patches around $N$ pixels in the resulting alpha matte, as well as over all possible combinations of pixel pairs $i,j$ in a given $3 \times 3$ patch, where the total number of combinations is $(3^2)*(3^2)=81$. The weights $w_{i,j}^n$ are given in \ref{eq:matting_laplacian_color}. We can then add the tensorized data term to get the final loss function:
\begin{equation}
\begin{split}
  E(\bX,\balpha) &= \sum_{b=1}^B \sum_{n=1}^N \sum_{k=1}^K \bW \odot (\tilde{\balpha}_I - \tilde{\balpha}_J)^2  \\
                 &+ \lambda \sum_{b=1}^B \sum_{n=1}^N \left[\bX_F + \bX_B\right] \odot \left[\balpha-\bX_F\right]^2, \\
\end{split}
\end{equation}
where $k \in [1...K=81]$ indexes the pixel pairs $i,j$ in a $3 \times 3$ patch, and $\bW \in \mathbb{R}^{B \times N \times 81}$ is the matrix of weights. $\tilde{\balpha}_I,\tilde{\balpha}_J \in \mathbb{R}^{B \times N \times 81}$ are repetitions of the alpha matte; the first represents the alpha mattes in index $i \rightarrow (1,..,1,2,...,2,...,9,...,9) \in \mathbb{R}^{81}$, and the second represents the alpha mattes in index $j \rightarrow (1,2,...,9,1,2,...,9,...,1,2,...9) \in \mathbb{R}^{81}$. The data term is exactly the same as in seeded segmentation, only there is no need for summation over the classes $l$; $\bX_F \in \mathbb{R}^{B \times N}$ and $\bX_B \in \mathbb{R}^{B \times N}$ are the foreground and background seed images.

\smallpar{Single Image Dehazing}
Note that the energy for this task, given in Equation \ref{eq:dehazing_energy}, is exactly identical to the image matting energy in \ref{eq:matting_energy}; $\balpha,\bX$ are replaced with $\bt,\tilde{\bt}$ correspondingly, and the seed matrix $\bQ$ is now identity. Following the same steps as in matting tensorization, we get the following loss function:
\begin{equation}
  E(\bt,\tilde{\bt}) = \sum_{b=1}^B \sum_{n=1}^N \sum_{k=1}^K \bW \odot (\bT_I - \bT_J)^2  
                 + \lambda \sum_{b=1}^B \sum_{n=1}^N (\bt - \tilde{\bt})^2,
\end{equation}
where $\bT_I,\bT_J$ are constructed as $\tilde{\balpha}_I,
\tilde{\balpha}_J$ in image matting. The coarse transmission map $\tilde{\bt}$ is calculated using \ref{eq:dark_channel_prior}.

%


\ifCLASSOPTIONcaptionsoff
  \newpage
\fi

\bibliographystyle{ieeetr}
\bibliography{IEEEabrv,bib_for_NIPS}

\begin{thebibliography}{10}

\bibitem{deep_learning}
Y.~LeCun, Y.~Bengio, and G.~Hinton, ``Deep learning,'' {\em nature}, vol.~521,
  no.~7553, p.~436, 2015.

\bibitem{speech_recognition}
D.~Amodei, S.~Ananthanarayanan, R.~Anubhai, J.~Bai, E.~Battenberg, C.~Case,
  J.~Casper, B.~Catanzaro, Q.~Cheng, G.~Chen, {\em et~al.}, ``Deep speech 2:
  End-to-end speech recognition in english and mandarin,'' in {\em
  International Conference on Machine Learning}, pp.~173--182, 2016.

\bibitem{fast_image_processing}
Q.~Chen, J.~Xu, and V.~Koltun, ``Fast image processing with fully-convolutional
  networks,'' in {\em IEEE International Conference on Computer Vision},
  vol.~9, 2017.

\bibitem{alexnet}
Y.~Jia, E.~Shelhamer, J.~Donahue, S.~Karayev, J.~Long, R.~Girshick,
  S.~Guadarrama, and T.~Darrell, ``Caffe: Convolutional architecture for fast
  feature embedding,'' in {\em Proceedings of the 22nd ACM international
  conference on Multimedia}, pp.~675--678, ACM, 2014.

\bibitem{NLP}
X.~Zhang, J.~Zhao, and Y.~LeCun, ``Character-level convolutional networks for
  text classification,'' in {\em Advances in neural information processing
  systems}, pp.~649--657, 2015.

\bibitem{whats_the_point}
A.~Bearman, O.~Russakovsky, V.~Ferrari, and L.~Fei-Fei, ``What’s the point:
  Semantic segmentation with point supervision,'' in {\em European Conference
  on Computer Vision}, pp.~549--565, Springer, 2016.

\bibitem{medical_1}
F.~Milletari, N.~Navab, and S.-A. Ahmadi, ``V-net: Fully convolutional neural
  networks for volumetric medical image segmentation,'' in {\em 2016 Fourth
  International Conference on 3D Vision (3DV)}, pp.~565--571, IEEE, 2016.

\bibitem{medical_2}
Q.~Dou, L.~Yu, H.~Chen, Y.~Jin, X.~Yang, J.~Qin, and P.-A. Heng, ``3d deeply
  supervised network for automated segmentation of volumetric medical images,''
  {\em Medical image analysis}, vol.~41, pp.~40--54, 2017.

\bibitem{CAP}
Q.~Zhu, J.~Mai, L.~Shao, {\em et~al.}, ``A fast single image haze removal
  algorithm using color attenuation prior.,'' {\em TIP}, vol.~24, no.~11,
  pp.~3522--3533, 2015.

\bibitem{mscnn}
W.~Ren, S.~Liu, H.~Zhang, J.~Pan, X.~Cao, and M.-H. Yang, ``Single image
  dehazing via multi-scale convolutional neural networks,'' in {\em ECCV},
  2016.

\bibitem{dehazenet}
B.~Cai, X.~Xu, K.~Jia, C.~Qing, and D.~Tao, ``Dehazenet: An end-to-end system
  for single image haze removal,'' {\em TIP}, vol.~25, no.~11, pp.~5187--5198,
  2016.

\bibitem{aodnet}
B.~Li, X.~Peng, Z.~Wang, J.~Xu, and D.~Feng, ``Aod-net: All-in-one dehazing
  network,'' in {\em ICCV}, 2017.

\bibitem{GFN}
W.~Ren, L.~Ma, J.~Zhang, J.~Pan, X.~Cao, W.~Liu, and M.-H. Yang, ``Gated fusion
  network for single image dehazing,'' {\em CVPR}, 2018.

\bibitem{energy_stereo}
Y.~Boykov and V.~Kolmogorov, ``An experimental comparison of min-cut/max-flow
  algorithms for energy minimization in vision,'' {\em IEEE transactions on
  pattern analysis and machine intelligence}, vol.~26, no.~9, pp.~1124--1137,
  2004.

\bibitem{energy_sr}
M.~Elad and A.~Feuer, ``Restoration of a single superresolution image from
  several blurred, noisy, and undersampled measured images,'' {\em IEEE
  transactions on image processing}, vol.~6, no.~12, pp.~1646--1658, 1997.

\bibitem{energy_dehazing}
K.~He, J.~Sun, and X.~Tang, ``Single image haze removal using dark channel
  prior,'' {\em IEEE transactions on pattern analysis and machine
  intelligence}, vol.~33, no.~12, pp.~2341--2353, 2011.

\bibitem{energy_optic}
A.~Wedel, D.~Cremers, T.~Pock, and H.~Bischof, ``Structure-and motion-adaptive
  regularization for high accuracy optic flow,'' in {\em Computer Vision, 2009
  IEEE 12th International Conference on}, pp.~1663--1668, IEEE, 2009.

\bibitem{deep_image_prior}
D.~Ulyanov, A.~Vedaldi, and V.~Lempitsky, ``Deep image prior,'' {\em arXiv
  preprint arXiv:1711.10925}, 2017.

\bibitem{style_transfer}
L.~A. Gatys, A.~S. Ecker, and M.~Bethge, ``A neural algorithm of artistic
  style,'' {\em arXiv preprint arXiv:1508.06576}, 2015.

\bibitem{FF_style_transfer}
J.~Johnson, A.~Alahi, and L.~Fei-Fei, ``Perceptual losses for real-time style
  transfer and super-resolution,'' in {\em European Conference on Computer
  Vision}, pp.~694--711, Springer, 2016.

\bibitem{FF_texture}
D.~Ulyanov, V.~Lebedev, A.~Vedaldi, and V.~S. Lempitsky, ``Texture networks:
  Feed-forward synthesis of textures and stylized images.,'' in {\em ICML},
  pp.~1349--1357, 2016.

\bibitem{unsupervised_optical_flow}
Z.~Ren, J.~Yan, B.~Ni, B.~Liu, X.~Yang, and H.~Zha, ``Unsupervised deep
  learning for optical flow estimation.,'' in {\em AAAI}, pp.~1495--1501, 2017.

\bibitem{unsupervised_face_reconstruction}
E.~Richardson, M.~Sela, R.~Or-El, and R.~Kimmel, ``Learning detailed face
  reconstruction from a single image,'' in {\em Computer Vision and Pattern
  Recognition (CVPR), 2017 IEEE Conference on}, pp.~5553--5562, IEEE, 2017.

\bibitem{unsupervised_smoothing}
Q.~Fan, J.~Yang, D.~Wipf, B.~Chen, and X.~Tong, ``Image smoothing via
  unsupervised learning,'' in {\em SIGGRAPH Asia 2018 Technical Papers},
  p.~259, ACM, 2018.

\bibitem{normalized_cut_energy}
M.~Tang, A.~Djelouah, F.~Perazzi, Y.~Boykov, and C.~Schroers, ``Normalized cut
  loss for weakly-supervised cnn segmentation,'' in {\em IEEE conference on
  Computer Vision and Pattern Recognition (CVPR), Salt Lake City}, 2018.

\bibitem{classical_seg_1}
C.~Rother, V.~Kolmogorov, and A.~Blake, ``Grabcut: Interactive foreground
  extraction using iterated graph cuts,'' in {\em ACM transactions on graphics
  (TOG)}, vol.~23, pp.~309--314, ACM, 2004.

\bibitem{classical_seg_2}
Y.~Y. Boykov and M.-P. Jolly, ``Interactive graph cuts for optimal boundary \&
  region segmentation of objects in nd images,'' in {\em Computer Vision, 2001.
  ICCV 2001. Proceedings. Eighth IEEE International Conference on}, vol.~1,
  pp.~105--112, IEEE, 2001.

\bibitem{random_walker}
L.~Grady, ``Random walks for image segmentation,'' {\em IEEE transactions on
  pattern analysis and machine intelligence}, vol.~28, no.~11, pp.~1768--1783,
  2006.

\bibitem{pascal}
M.~Everingham, L.~Van~Gool, C.~K. Williams, J.~Winn, and A.~Zisserman, ``The
  pascal visual object classes (voc) challenge,'' {\em International journal of
  computer vision}, vol.~88, no.~2, pp.~303--338, 2010.

\bibitem{imagenet}
O.~Russakovsky, J.~Deng, H.~Su, J.~Krause, S.~Satheesh, S.~Ma, Z.~Huang,
  A.~Karpathy, A.~Khosla, M.~Bernstein, {\em et~al.}, ``Imagenet large scale
  visual recognition challenge,'' {\em International Journal of Computer
  Vision}, vol.~115, no.~3, pp.~211--252, 2015.

\bibitem{supervised_segmentation}
J.~Long, E.~Shelhamer, and T.~Darrell, ``Fully convolutional networks for
  semantic segmentation,'' in {\em Proceedings of the IEEE conference on
  computer vision and pattern recognition}, pp.~3431--3440, 2015.

\bibitem{weakly_darrell}
D.~Pathak, P.~Krahenbuhl, and T.~Darrell, ``Constrained convolutional neural
  networks for weakly supervised segmentation,'' in {\em Proceedings of the
  IEEE international conference on computer vision}, pp.~1796--1804, 2015.

\bibitem{seed_expand}
A.~Kolesnikov and C.~H. Lampert, ``Seed, expand and constrain: Three principles
  for weakly-supervised image segmentation,'' in {\em European Conference on
  Computer Vision}, pp.~695--711, Springer, 2016.

\bibitem{stc}
Y.~Wei, X.~Liang, Y.~Chen, X.~Shen, M.-M. Cheng, J.~Feng, Y.~Zhao, and S.~Yan,
  ``Stc: A simple to complex framework for weakly-supervised semantic
  segmentation,'' {\em IEEE transactions on pattern analysis and machine
  intelligence}, vol.~39, no.~11, pp.~2314--2320, 2017.

\bibitem{weaklysup}
G.~Papandreou, L.-C. Chen, K.~P. Murphy, and A.~L. Yuille, ``Weakly-and
  semi-supervised learning of a deep convolutional network for semantic image
  segmentation,'' in {\em Proceedings of the IEEE international conference on
  computer vision}, pp.~1742--1750, 2015.

\bibitem{boxsup}
J.~Dai, K.~He, and J.~Sun, ``Boxsup: Exploiting bounding boxes to supervise
  convolutional networks for semantic segmentation,'' in {\em Proceedings of
  the IEEE International Conference on Computer Vision}, pp.~1635--1643, 2015.

\bibitem{scribblesup}
D.~Lin, J.~Dai, J.~Jia, K.~He, and J.~Sun, ``Scribblesup: Scribble-supervised
  convolutional networks for semantic segmentation,'' in {\em Proceedings of
  the IEEE Conference on Computer Vision and Pattern Recognition},
  pp.~3159--3167, 2016.

\bibitem{closed_form_matting}
A.~Levin, D.~Lischinski, and Y.~Weiss, ``A closed-form solution to natural
  image matting,'' {\em IEEE Transactions on Pattern Analysis and Machine
  Intelligence}, vol.~30, no.~2, pp.~228--242, 2008.

\bibitem{classic_matting_1}
Y.-Y. Chuang, B.~Curless, D.~H. Salesin, and R.~Szeliski, ``A bayesian approach
  to digital matting,'' in {\em Computer Vision and Pattern Recognition, 2001.
  CVPR 2001. Proceedings of the 2001 IEEE Computer Society Conference on},
  vol.~2, pp.~II--II, IEEE, 2001.

\bibitem{classic_matting_2}
E.~S. Gastal and M.~M. Oliveira, ``Shared sampling for real-time alpha
  matting,'' in {\em Computer Graphics Forum}, vol.~29, pp.~575--584, Wiley
  Online Library, 2010.

\bibitem{classic_matting_3}
K.~He, C.~Rhemann, C.~Rother, X.~Tang, and J.~Sun, ``A global sampling method
  for alpha matting,'' in {\em Computer Vision and Pattern Recognition (CVPR),
  2011 IEEE Conference on}, pp.~2049--2056, IEEE, 2011.

\bibitem{classic_matting_4}
L.~Grady, T.~Schiwietz, S.~Aharon, and R.~Westermann, ``Random walks for
  interactive alpha-matting,'' in {\em Proceedings of VIIP}, vol.~2005,
  pp.~423--429, 2005.

\bibitem{classic_matting_5}
J.~Wang and M.~F. Cohen, ``Optimized color sampling for robust matting,'' in
  {\em Computer Vision and Pattern Recognition, 2007. CVPR'07. IEEE Conference
  on}, pp.~1--8, IEEE, 2007.

\bibitem{deep_matting}
N.~Xu, B.~Price, S.~Cohen, and T.~Huang, ``Deep image matting,'' in {\em
  Computer Vision and Pattern Recognition (CVPR)}, 2017.

\bibitem{deep_matting_1}
X.~Shen, X.~Tao, H.~Gao, C.~Zhou, and J.~Jia, ``Deep automatic portrait
  matting,'' in {\em European Conference on Computer Vision}, pp.~92--107,
  Springer, 2016.

\bibitem{deep_matting_2}
D.~Cho, Y.-W. Tai, and I.~Kweon, ``Natural image matting using deep
  convolutional neural networks,'' in {\em European Conference on Computer
  Vision}, pp.~626--643, Springer, 2016.

\bibitem{alpha_matting_com}
C.~Rhemann, C.~Rother, J.~Wang, M.~Gelautz, P.~Kohli, and P.~Rott, ``A
  perceptually motivated online benchmark for image matting,'' in {\em Computer
  Vision and Pattern Recognition, 2009. CVPR 2009. IEEE Conference on},
  pp.~1826--1833, IEEE, 2009.

\bibitem{contrast_tan}
R.~T. Tan, ``Visibility in bad weather from a single image,'' in {\em CVPR},
  2008.

\bibitem{BCCR}
G.~Meng, Y.~Wang, J.~Duan, S.~Xiang, and C.~Pan, ``Efficient image dehazing
  with boundary constraint and contextual regularization,'' in {\em ICCV},
  2013.

\bibitem{color_lines}
R.~Fattal, ``Dehazing using color-lines,'' {\em TOG}, vol.~34, no.~1, p.~13,
  2014.

\bibitem{NLD}
D.~Berman, T.~Tali, and S.~Avidan, ``Non-local image dehazing,'' in {\em CVPR},
  2016.

\bibitem{haze_model}
W.~K. Middleton, ``Vision through the atmosphere,'' in {\em Geophysik
  II/Geophysics II}, pp.~254--287, Springer, 1957.

\bibitem{reside}
B.~Li, W.~Ren, D.~Fu, D.~Tao, D.~Feng, W.~Zeng, and Z.~Wang, ``Benchmarking
  single-image dehazing and beyond,'' {\em IEEE Transactions on Image
  Processing}, vol.~28, no.~1, pp.~492--505, 2019.

\bibitem{optical_flow}
D.~Fleet and Y.~Weiss, ``Optical flow estimation,'' in {\em Handbook of
  mathematical models in computer vision}, pp.~237--257, Springer, 2006.

\bibitem{depth_estimate}
A.~Levin, R.~Fergus, F.~Durand, and W.~T. Freeman, ``Image and depth from a
  conventional camera with a coded aperture,'' {\em ACM transactions on
  graphics (TOG)}, vol.~26, no.~3, p.~70, 2007.

\bibitem{retargetting}
M.~Rubinstein, D.~Gutierrez, O.~Sorkine, and A.~Shamir, ``A comparative study
  of image retargeting,'' in {\em ACM transactions on graphics (TOG)}, vol.~29,
  p.~160, ACM, 2010.

\bibitem{retinex}
D.~J. Jobson, Z.-u. Rahman, and G.~A. Woodell, ``A multiscale retinex for
  bridging the gap between color images and the human observation of scenes,''
  {\em IEEE Transactions on Image processing}, vol.~6, no.~7, pp.~965--976,
  1997.

\bibitem{medical_seeded_seg}
C.~Couprie, L.~Najman, and H.~Talbot, ``Seeded segmentation methods for medical
  image analysis,'' in {\em Medical Image Processing}, pp.~27--57, Springer,
  2011.

\bibitem{CAN}
F.~Yu and V.~Koltun, ``Multi-scale context aggregation by dilated
  convolutions,'' {\em arXiv preprint arXiv:1511.07122}, 2015.

\bibitem{Resnet}
K.~He, X.~Zhang, S.~Ren, and J.~Sun, ``Deep residual learning for image
  recognition,'' in {\em Proceedings of the IEEE conference on computer vision
  and pattern recognition}, pp.~770--778, 2016.

\bibitem{pascal_aug}
B.~Hariharan, P.~Arbel{\'a}ez, L.~Bourdev, S.~Maji, and J.~Malik, ``Semantic
  contours from inverse detectors,'' in {\em 2011 International Conference on
  Computer Vision}, pp.~991--998, IEEE, 2011.

\bibitem{Adam}
D.~P. Kingma and J.~Ba, ``Adam: A method for stochastic optimization,'' {\em
  arXiv preprint arXiv:1412.6980}, 2014.

\bibitem{implicit_bias}
S.~Gunasekar, J.~Lee, D.~Soudry, and N.~Srebro, ``Implicit bias of gradient
  descent on linear convolutional networks,'' {\em arXiv preprint
  arXiv:1806.00468}, 2018.

\end{thebibliography}

\begin{IEEEbiography}[{\includegraphics[width=1in,height=1.25in,clip,keepaspectratio]{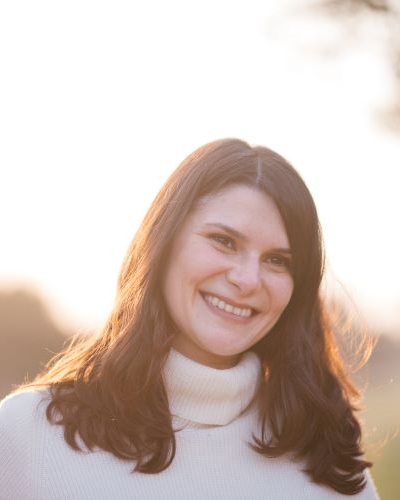}}]{Alona Golts}
received the B.Sc. and M.Sc. degrees from the Department of Electrical Engineering, Technion—Israel Institute of Technology, Haifa, Israel, in 2010 and 2015, respectively. She is
currently pursuing her Ph.D. in the department of Computer Science in the Technion. Her research interests are deep learning, inverse problems and sparse representations.
\end{IEEEbiography}

\vskip -20pt plus -1fil

\begin{IEEEbiography}[{\includegraphics[width=1in,height=1.25in,clip,keepaspectratio]{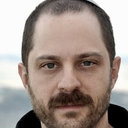}}]{Daniel Freedman}
received the AB in Physics from Princeton University (Magna Cum Laude) in 1993, and his Ph.D. in Engineering Sciences from Harvard University in 2000. From 2000-9, he served as Assistant Professor and Associate Professor in the Computer Science Department at Rensselaer Polytechnic Institute (RPI) (Troy, NY). In 2007, he became a Fulbright Fellow and Visiting Professor of Applied Mathematics and Computer Science at the Weizmann Institute of Science. He then worked in a number of research positions in HP Labs, IBM Research, Microsoft Research, and finally Google Research. In addition to the Fulbright Fellowship, he received the National Science Foundation CAREER Award.
\end{IEEEbiography}

\vskip -20pt plus -1fil


\begin{IEEEbiography}[{\includegraphics[width=1in,height=1.25in,clip,keepaspectratio]{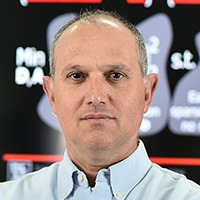}}]{Michael Elad}
received the B.Sc., M.Sc., and
D.Sc. degrees from the Department of Electrical engineering, Technion—Israel Institute of Technology,
Haifa, Israel, in 1986, 1988, and 1997, respectively.
Since 2003, he has been a faculty member in the
Department of Computer Science, Technion—Israel
Institute of Technology, where since 2010, he is a
Full Professor. He works in the field of signal and
image processing, specializing in inverse problems,
and sparse representations. He was the recipient of
numerous teaching awards, the 2008 and 2015 Henri
Taub Prizes for Academic Excellence, and the 2010 Hershel-Rich prize for innovation. He is a SIAM Fellow (2018). Since January 2016, He has been the
Editor-in-Chief for SIAM Journal on Imaging Sciences since January 2016.
\end{IEEEbiography}




\end{document}